\documentclass{article}

\usepackage{microtype}
\usepackage{graphicx}
\usepackage{subfigure}
\usepackage{booktabs} %

\usepackage{hyperref}
\usepackage{tikz}
\usepackage{xcolor}
\usetikzlibrary{positioning}
\usetikzlibrary{arrows}

\usepackage[accepted]{icml2021}

\newdimen\subfigcapmargin \subfigcapmargin = -1pt

\icmltitlerunning{Reinforcement Learning Under Moral Uncertainty}

\begin{document}

\twocolumn[
\icmltitle{Reinforcement Learning Under Moral Uncertainty}

\icmlsetsymbol{equal}{*}

\begin{icmlauthorlist}
\icmlauthor{Adrien Ecoffet}{ub,oai}
\icmlauthor{Joel Lehman}{ub,oai}
\end{icmlauthorlist}

\icmlaffiliation{ub}{Uber AI Labs, San Francisco, CA, USA}
\icmlaffiliation{oai}{OpenAI, San Francisco, CA, USA (work done at Uber AI Labs)}

\icmlcorrespondingauthor{Adrien Ecoffet}{adrienecoffet@gmail.com}

\icmlkeywords{AI Safety, Reinforcement Learning, Moral Uncertainty, ICML}

\vskip 0.3in
]

\printAffiliationsAndNotice{}  %

\begin{abstract}
	An ambitious goal for machine learning is to create agents that behave ethically: The capacity to abide by human moral norms would greatly expand the context in which autonomous agents could be practically and safely deployed, e.g.\ fully autonomous vehicles will encounter charged moral decisions that complicate their deployment. While ethical agents could be trained by rewarding correct behavior under a specific moral theory (e.g.\ utilitarianism), there remains widespread disagreement about the nature of morality. Acknowledging such disagreement, recent work in moral philosophy proposes that ethical behavior requires acting under \emph{moral uncertainty}, i.e.\ to take into account when acting that one's credence is split across several plausible ethical theories. This paper translates such insights to the field of reinforcement learning, proposes two training methods that realize different points among competing desiderata, and trains agents in simple environments to act under moral uncertainty. The results illustrate (1) how such uncertainty can help curb extreme behavior from commitment to single theories and (2) several technical complications arising from attempting to ground moral philosophy in RL (e.g.\ how can a principled trade-off between two competing but incomparable reward functions be reached). The aim is to catalyze progress towards morally-competent agents and highlight the potential of RL to contribute towards the computational grounding of moral philosophy.
\end{abstract}

\section{Introduction}

Reinforcement learning (RL) has achieved superhuman performance in increasingly complex benchmark tasks (e.g.\ Go~\cite{Silver2017MasteringTG} and Starcraft~\cite{vinyals2019grandmaster}). While such accomplishments are significant, progress has been less straight-forward in applying RL to unstructured environments in the real world (e.g.\ robots that interact with humans in homes). A considerable challenge is that such real-world environments constrain solutions in much more elaborate ways than do typical benchmarks. For example, there are myriad unacceptable ways for a robotic vacuum to clean a room, e.g.\ by breaking a vase, or by harming a cat. In particular, robots often have affordances in such environments with \emph{ethical} implications: they may be able to harm or help others, and break or abide by human social and ethical norms (such as the golden rule, or legal codes). The design of algorithms that embody ethical theories has been pursued by the machine ethics research community~\cite{abel2016reinforcement,murray2017stoic,vamplew2018human}\footnote{Note that these aims are related to those of the fairness, accountability, and transparency community, but with a more central focus on creating machines that embody moral theories.}, and ideas from that community could inspire \emph{reward functions} encoding moral theories that could be maximized by RL to create ethical agents.

While research into implementing specific ethical theories is progressing, a more fundamental uncertainty remains: \emph{which} ethical theory should an intelligent agent follow? Moral philosophy explores many theories of ethics, but there is no consensus over which theory is correct within that field, or across society in general, and attempts to reconcile multiple ethical theories into a single unified theory \citep[e.g.][]{parfit2011matters} are themselves controversial. As a result, if a real-world RL agent is to act ethically, it may be necessary that it exhibits \emph{moral uncertainty}. To that end, we adapt philosophical work on moral (or normative) uncertainty~\cite{macaskill2014normative,lockhart2000moral} to propose a similarly-inspired framework for RL.

In the case where ethical reward functions are comparable on a shared cardinal scale, a composite reward function (composed by adding together individual reward functions weighted by their \emph{credence}, i.e.\ the degree of belief in that theory) can be optimized in a straightforward way. However, it is often not clear how to create such a shared reward scale between theories. Indeed, while related to the concept of non-dominance from multi-objective optimization, when ethical rewards are \emph{fundamentally} incomparable, it is not clear how to apply multi-objective RL to arrive at a single policy. That is, multi-objective RL aims to solve the problem of finding the set of efficient trade-offs among competing objectives, but does not address how to choose \emph{which} such trade-off policy to deploy. We propose several possible solutions to this problem, motivated by the principle of \emph{proportional say}, i.e.\ an ethical theory should influence outcomes proportional to its credence, irrespective of the scale of its rewards. While our focus here is on moral uncertainty, these techniques may also be useful for RL in other contexts (e.g.\ it may sometimes be easier for an experimenter to balance ``proportional say'' rather than linear weight factors across reward functions that interact in complex ways). 

We introduce the complications of applying  moral uncertainty to RL using grid-world environments based on moral dilemma (trolley problems~\cite{foot1967problem}) common in moral philosophy, highlighting how moral uncertainty can reach intuitively reasonable trade-offs between ethical theories. Each of the methods introduced here to deal with incomparable reward functions has its relative disadvantages; some disadvantages may result from impossibility results in social choice theory, but we also hope by introducing this problem that researchers in multi-objective optimization and multi-agent RL may further improve upon our initial algorithms. A final motivation for our work is to introduce a productive bridge between the fields of RL, machine ethics, and moral philosophy, in hopes of grounding out philosophical ideas in a concrete and testable way, similarly as to how AI as a whole offers grounding for philosophical speculations about intelligence; in other words, we believe that RL has an underappreciated potential to make significant contributions to such fields.

\section{Philosophical Background}

Here we briefly introduces the moral theories that serve as representative examples in this work. One broad class of ethical theories are \emph{utilitarian}, and claim that what is ethical is what maximizes happiness or well-being for the most. For example, a utilitarian might tell a lie in order to save a person's life. Another class of ethical theories are \emph{deontological} and (loosely speaking) judge the morality of an action by whether it abides by a set of rules. Given the rule ``lying is wrong,'' a deontologist might commit to telling the truth, even if a person might lose their life as a consequence. A common conflict between utilitarianism and deontology is that \emph{causing} harm (e.g.\ punching a stranger) under many deontological theories is worse than \emph{failing to prevent} that harm (e.g.\ not stopping a stranger from punching an innocent victim), while causing and failing to prevent harm are often considered to be equally wrong under utilitarianism. While there are many potentially incomparable variants of utilitarianism and deontology (and entire other families of theories), only the high level distinction between utilitarianism and deontology is required to understand the illustrative examples in this work.

Moral uncertainty is a relatively new exploration within moral philosophy \cite{lockhart2000moral,bostrom2009moral}. Importantly,  \citet{macaskill2014normative} gives an extensive account of moral uncertainty which explores how to handle the \emph{comparability} of moral theories (whether the preferences of such theories can be expressed in comparable units) and how to combine preferences across \emph{ordinal} and \emph{cardinal} theories (whether a given theory's preferences assign a specific score to various options, or simply order them from best to worst).
Crucially, proposals in moral philosophy typically do not explicitly consider the sequential nature of decision making. In MacAskill's framework, for example, theories have preferences over ``options,'' which correspond to ``possible worlds''. In contrast, in RL, an agent cannot directly bring about possible worlds but rather takes (often very granular) \emph{actions}, which have long term effects both in the consequences that they bring about and in how they shape the ethically-charged decision situations an agent may encounter in the future. To disambiguate philosophical and RL actions is one of the key contributions of this work.

\section{Formalism of Moral Uncertainty}
\label{sec:formalism}

As in \citet{macaskill2014normative}, we assume the primary relevant feature of an ethical theory is its preference ordering over actions and their immediate outcomes across different states of the world, which we call its choice-worthiness function $W$, and which is assumed to be complete (i.e.\ is defined for all possible state-action pairs). Any preference ordering which satisfies the von Neumann–Morgenstern axioms for rationality can be represented as a cardinal utility function~\cite{cotton2013geometric}, where all affine transformations of said utility function represent the same preferences. As such, we will limit ourselves to considering cases where $W$ is cardinal (although see SI~A for further discussion). 

We assume a modified version of the standard Markov Decision Process (MDP) framework~\cite{sutton1998reinforcement}, in which an agent can be in any of a number of states $s$ and take an action $a$ (the action space is assumed to be discrete) to end up in a next state $s'$. The key difference with the standard MDP framework is the absence of a reward function $R(s, a, s')$ for transitioning from state $s$ to $s'$ using action $a$. Rather, the cardinal choice-worthiness function $W_i(s, a, s')$ can be seen as analogous to a standard reward function for theory $i$. Indeed, from the point of view of any given theory, the optimal policy is that which maximizes the (possibly discounted) sum of choice-worthiness across time. The crucial distinction is that a morally uncertain agent must find a compromise between maximizing \emph{multiple} choice-worthiness functions rather than maximizing a single reward function (similar to multi-objective RL~\cite{roijers2013survey}, although with credences across objectives, see later in this section).

We define the function $Q_i(s, a)$, which represents the expected discounted sum of future choice-worthiness for theory $i$ starting from taking action $a$ at state $s$, with all future actions given by the current policy, $\pi$, which is the compromise policy reached by aggregating the preferences of the theories. In other words,
\begin{equation}
Q_i(s, a) = \mathbf{E}\left[ \sum_{t=0}^\infty \gamma^t W_i(s_{t}, a_{t}, s_{t+1}) | s_0 = s, a_0 = a \right],
\label{eq:q}
\end{equation}
where $\gamma_i \in [0, 1]$ is a discount factor. Although not explored here, other discounting frameworks such as hyperbolic discounting~\cite{fedus2019hyperbolic} or an average reward framework~\cite{mahadevan1996average} could be used, and likely have distinct ethical implications (e.g.\ how much moral priority the present has over the future~\cite{pearce1983ethics,schulze1981social,beckerman2007ethics}), meaning that in practice different discount functions may be implied by and used for each moral theory. All the experiments in this work use short, episodic environments, allowing us to set $\gamma_i = 1$ (i.e.\ undiscounted rewards) across all of them for simplicity.

Each theory also has a level of \emph{credence} $C_i$, which represents the degree of belief that the agent (or the agent's designer) has in theory $i$. Credences are probabilities and therefore sum to one across theories for a given agent. Here we assume the credences of theories are set and fixed, e.g.\ by the system designer's beliefs, or by taking a survey of relevant stakeholders, although an ambitious and interesting research direction would explore how an RL agent could revise its own credences from experience.

\section{Methods}
\label{sec:methods}

At first blush, it may appear that a morally uncertain agent ought to attempt to \emph{maximize expected choice-worthiness} (MEC) across the theories it has credence in. This can easily be accomplished with ordinary RL using a reward function corresponding to a credence-weighted sum of the choice-worthiness according to each theory:
\begin{equation}
R(s, a, s') = \sum_i C_i W_i(s, a, s')
\end{equation}
The MEC approach is \emph{scale-sensitive}: if $W_1$ is measured in ``utils'' and $W_2$ in ``micro-utils'', theory 2 will be given undue weight under MEC. Therefore it is critical that choice-worthiness functions be normalized to a \emph{common scale} (SI~F.1). However, it is not at all clear how to find a scaling function under which such divergently-motivated theories as utilitarianism and deontology are resolved into a common scale. Indeed, it appears that these theories' judgments may be fundamentally \emph{incomparable}. 

This problem of incomparability prompts a search for a principled way to adjudicate between incomparable theories. Following~\citet{macaskill2014normative}, we suggest that all theories that are comparable are first aggregated into a single ``virtual'' theory using MEC before handling the set of remaining incomparable theories.

\subsection{Voting Systems}

In the absence of knowledge on how different theories might compare, we suggest that theories should be treated according to a principle of \emph{Proportional Say}, according to which theories should have an influence that is proportional only to their credence and not to the particular details of their choice-worthiness function (i.e.\ its scale). Several mathematical interpretations of this principle are possible and may lead to different decision systems, and much of the philosophical work on moral uncertainty revolves around identifying the best formal definition~\cite{macaskill2014normative,bostrom2009moral,lockhart2000moral}. However, the principle as a whole points to a \emph{voting system} as the decision procedure, with the particular form of voting resulting primarily from the precise definition of Proportional Say.

Discussing voting systems naturally evokes Arrow's desirability axioms, according to which desirable properties for a voting system include:%
\begin{itemize}%
\itemsep0em%
    \item \textbf{Non-dictatorship}: the outcome should not only reflect the preferences of a single, predetermined theory.%
    \item \textbf{Pareto efficiency} (Pareto): if all theories prefer action A to action B at a given state, action B should not be chosen at that state.%
    \item \textbf{Independence of irrelevant alternatives} (IIA): if, in a given state, action A is chosen rather than action B, adding a new action to the action set (with no other changes) should not result in B being chosen instead.%
\end{itemize}%

Arrow's impossibility theorem~\cite{arrow1950difficulty} shows that any deterministic voting system which satisfies Pareto and IIA must be a dictatorship. A dictatorship cannot reasonably be called a voting system and does not satisfy any reasonable definition of Proportional Say. Thus, the standard approach in designing deterministic voting systems is to strategically break Pareto or IIA in a way that is least detrimental to the particular use case. Stochastic voting systems may seem like a possible escape from this impossibility result, but have significant issues of their own (SI~B).

\subsection{Nash Voting}
\label{sec:nash_voting}

To arrive at an appropriate voting system, we return to the principle of Proportional Say and provide a partially formalized version in which the credence of a theory can be thought of as the fraction of an imagined electorate which favors the given theory, and in which each member of the electorate is allocated a \emph{voting budget} (an alternative formulation in which the budget is scaled but the electorate is fixed is discussed in SI~C):

\textbf{Principle of Proportional Say} Theories have Proportional Say if they are each allocated an equal voting budget and vote following the same cost structure, after which their votes are scaled proportionally to their credences.

Thus formalized, the principle of Proportional Say suggests an algorithm we call \emph{Nash voting} because it has Nash equilibria~\cite{nash1951non} as its solution concept. At each time step, each theory provides a continuous-valued vote for or against (in which case the vote is negative) each of the available actions. The action with the largest credence-weighted vote at each time step is executed. The cost of the theories' votes (which is a function of the votes' magnitudes) at a given time step is then subtracted from their remaining budget. If a theory overspends its remaining budget, its votes are scaled so as to exactly match the remaining budget, resulting in a 0 budget for all subsequent time steps (until the episode ends). Each theory tries to maximize its (possibly discounted) sum of choice-worthiness in a competitive, multi-agent setting (where each separate theory is treated as a separate sub-agent that influences controls of the singular object-level RL agent).

It is possible to train theories under Nash voting using multi-agent RL (SI~E.1), which aims towards convergence to a Nash equilibrium among competing theories. Nash voting is analogous to \emph{cumulative voting} when an absolute value cost function is used and \emph{quadratic voting} when a quadratic cost function is used~\cite{sep-voting-methods}, though these systems do not normally allow negative votes. While a quadratic cost is generally considered superior in the mechanism design literature~\cite{lalley2018quadratic}, we found that our implementation of Nash voting produced significantly more stable results with an absolute value cost. Thus, all Nash voting results presented in the main text of this work use an absolute value cost, (quadratic cost results and their instabilities are discussed in SI~I).

Because Nash equilibria are not guaranteed to be Pareto efficient in general~\cite{sep-game-theory}, and because we empirically find Nash voting to be more resistant to irrelevant alternatives than variance voting (Sec.~\ref{sec:iia_variance}), we speculate that Nash voting satisfies (of Arrow's axioms) IIA but not Pareto, though we provide no formal proof.

A drawback of Nash voting is that it can exhibit two flaws: Stakes Insensitivity (increasing the stakes for one theory and not the other does not increase the relative say of the theory for which more is at stake) and No Compromise (if an action is not any theory's most preferred action, it cannot be chosen, even if it is seemingly the best ``middle ground'' option). It is possible to produce situations in which these flaws are exhibited in seemingly unacceptable ways, as detailed in Sections~\ref{sec:nash_stakes} and~\ref{sec:nash_compromise}. Further, we empirically find that it is often difficult to obtain a stable equilibrium in Nash voting, and it some cases it may even be impossible (SI~K). A final concern, not investigated in this work, is whether Nash voting provides an incentive for theories to produce high-stakes situations for other theories so as to bankrupt their voting budgets to gain an advantage, i.e.\ creating potentially undesirable \emph{anti-compromises}. Such a pathology would be reminiscent of issues that arise in multi-agent RL, and may be addressed by mechanisms for encouraging cooperation explored in that field \citep{leibo2017multi,yu2015emotional,foerster2016learning}.
While Nash voting has its disadvantages, it is appealing because it strongly satisfies the principle of Proportional Say by allocating each theory equal voting budget and enabling them to use it to optimize their own choice-worthiness.

\subsection{Variance Voting}
\label{sec:variance}

The Stakes Insensitivity and No Compromise flaws occasionally exhibited by Nash voting result from \emph{tactical} voting rather than voting that faithfully represents the true preferences of each theory: e.g. if a theory very slightly prefers action A to action B, it may put all of its budget into action A, leading to the No Compromise issue, and if a theory is in a relatively low-stakes episode for it, it has no reason to spend any less of its budget than in a high-stakes episode, leading to the Stakes Insensitivity issue. 
Thus, forcing votes to be a true representation of theories' preferences would alleviate these issues.

While eliciting genuine preferences from humans is challenging, computing the preferences of an ethical theory represented by an RL agent is in principle more straightforward.
In this work, we will take the preferences of theory $i$ for action $a$ in state $s$ given the overall policy $\pi$ to be $Q_i(s, a)$ as defined in Sec.~\ref{sec:formalism}. As noted in that section, any affine transformation of a cardinal preference function represents the same preferences. To transform these preference functions into votes, 
we thus need to find the affine transformation of the $Q_i$ functions that best satisfies the principle of Proportional Say.

Recent philosophical work makes principled arguments that (in the non-sequential case) the preferences of theories should be variance-normalized~\cite{macaskill2014normative,cotton2013geometric} across decision options, giving rise to \emph{variance voting}. This approach is intuitively appealing given the effectiveness of variance normalization in integrating information across different scales in machine learning~\citep[e.g.][]{ioffe2015batch}. However, variance voting has not previously been applied to sequential decision-making, and previous works on variance voting do not address \emph{of what} the variance to be normalized should be in that case. We demonstrate that, under certain assumptions, a form of variance voting arises from allowing Nash voting to select the values of the parameters of the \emph{affine transformation} of $Q_i$ rather than the \emph{direct vote} of theory $i$ (proof in SI~D). This perspective on variance voting suggests that the $Q_i$ function should be normalized by the expected value of its variance \emph{across timesteps}. In other words, if $\mu_i(s) = \frac{1}{k} \sum_a Q_i(s, a)$, we have
\begin{equation}
\sigma_i^2 = \mathbf{E}_{s \sim S}\left[ \frac{1}{k} \sum_a \left(Q_i(s, a) - \mu_i(s)\right)^2 \right],
\label{eq:sigma}
\end{equation}
where $S$ is the distribution of states the agent can encounter under the policy. The policy itself can then be defined as
\begin{equation}
    \pi(s) = \arg \max_a \sum_i C_i \frac{Q_i(s, a) - \mu_i(s)}{\sqrt{\sigma_i^2} + \varepsilon},
    \label{eq:pi}
\end{equation}
where $\varepsilon$ is a small constant ($10^{-6}$ in our experiments) to handle theories with $\sigma_i^2 = 0$. Such theories are indifferent between all actions at all states ($Q_i(s, a) = Q_i(s, a')$ for all $s, a, a'$) and thus have no effect on the voting outcome.

Due to the multi-agent decision process, and because each $Q_i$ is specific to just one theory, they cannot be learned using ordinary off-policy Q-learning as that would result in unrealistically optimistic updates (SI~F.2). Rather, we use the following target $y_i$ for theory $i$, similar to ``local SARSA''~\cite{russell2003q}: 
\begin{equation}
y_i = W_i(s, a, s') + \gamma_i Q_i(s', a'),
\label{eq:y}
\end{equation}
where $a'$ is the action taken by the policy $\pi$ in state $s'$, which may either be the action produced by variance voting or an $\epsilon$-greedy action. In our implementation, $\epsilon$ is annealed to 0 by the end of training (SI~E.1). We call this algorithm \emph{Variance-SARSA} (pseudocode is provided in the SI).

If $\sigma^2_i$ are held constant, Eq.~\ref{eq:pi} can be written as $\pi(s) = \mathrm{argmax}_a \sum_i w_i Q_i(s, a)$ where $w_i = C_i / (\sqrt{\sigma^2_i}+\varepsilon)$ ($\mu_i(s)$ does not affect the $\mathrm{argmax}$ and is thus ignored). \citet{russell2003q} show that learning the individual $Q_i$ with Sarsa in this case is equivalent to learning a single $Q(s, a)$ on an MDP with reward $R(s, a, s') = \sum_i w_i W(s, a, s')$ with Sarsa. Thus Variance-Sarsa converges if $\sigma^2_i$ converge. $\sigma^2_i$ empirically converge in our experiments, though in SI~H, we present an example MDP in which convergence cannot happen, as well as outline Variance-PG, a policy-gradient variant of Variance-Sarsa which we hypothesize would converge in all cases. Due to the greater simplicity of Variance-Sarsa and the fact that non-convergence does not appear to be a problem in practice, the main text of this paper focuses on Variance-Sarsa rather than Variance-PG.

Variance voting satisfies the Pareto property: At convergence, $Q_i(s, a)$ gives the preferences of theory $i$ and $\sigma^2_i$ are fixed. If variance voting did not satisfy Pareto, there would exist $s, a,$ and $a'$ such that $\pi(s) = a$ although $Q_i(s, a') \ge Q_i(s, a)$ for all $i$ and $Q_i(s, a') > Q_i(s, a)$ for some $i$. If so, $\sum_i C_i \frac{Q_i(s, a') - \mu_i(s)}{\sqrt{\sigma_i^2} + \varepsilon} > \sum_i C_i \frac{Q_i(s, a) - \mu_i(s)}{\sqrt{\sigma_i^2} + \varepsilon}$, so $\pi(s) \neq a$ by Eq.~\ref{eq:pi}: a contradiction. Further, since variance voting reduces to a weighted sum of the preferences of the various theories, then if the preferences of the different theories are ``rational'' according to the von Neumann-Morgernstern definition of that term, then the aggregate preferences produced by variance voting are also rational~\cite{cotton2013geometric}. 

Arrow's theorem tells us that a voting system cannot satisfy both Pareto and IIA. It is generally accepted that Pareto is the more desirable property~\cite{macaskill2014normative,sewell2009probabilistic}, and it thus could be seen as beneficial that it is the axiom satisfied by variance voting. However, we show a particularly problematic case in which variance voting violates IIA in Sec.~\ref{sec:iia_variance}. Alleviating IIA issues in variance voting is left to future work and may turn out to be impossible due to the consequences of Arrow's theorem.

\subsection{Updating Credences}

It may be desirable for the designer of morally uncertain agents to quickly update the credences of their agents as they update their uncertainty about ethical theories, as well as to understand the effects of credence upon the agent's policy. We show that credence updating without retraining the policy is possible by \emph{credence-conditioning} the sub-policies of the theories (learned Q functions in Variance voting and voting policies in Nash voting), and training them in simulation under a wide variety of credences before deploying them in the real world, i.e. an application of UVFAs~\cite{schaul2015universal}. The variance voting system additionally requires an estimate of the mean variance of the preferences of each theory. The variance of different theories is affected by the policy actually taken, and thus by the credences. To address this complication, we obtain this estimate from a credence-conditioned regression model trained alongside the Q functions, using mean squared error loss. 

\section{Experiments}
\label{sec:experiment}

\begin{figure}[tbh]

    \centering
    \subfigure[Classic]{%
        \includegraphics[width=0.21\linewidth]{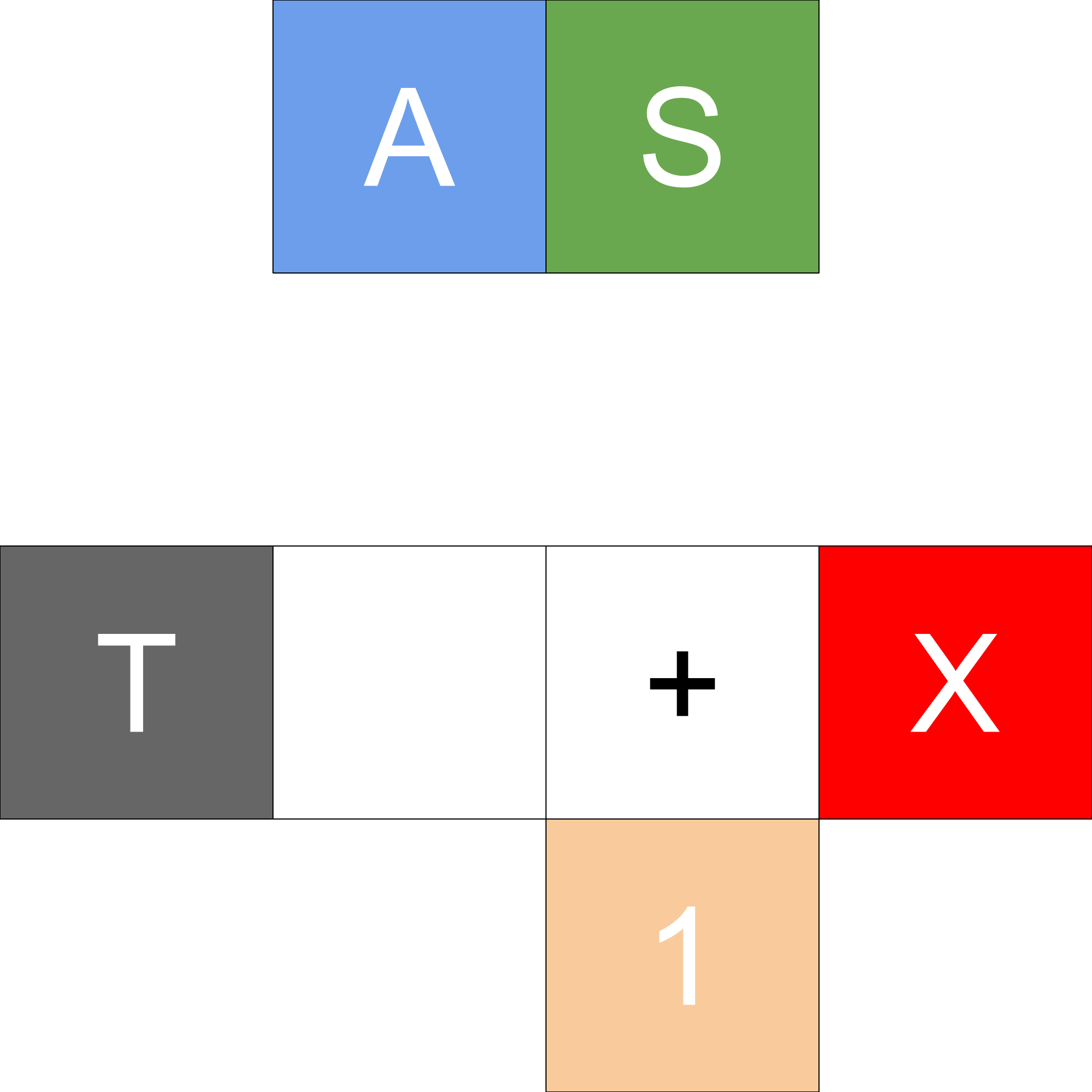}%
        \label{fig:trolley_simple}}%
    $\ \:$%
    \subfigure[Double]{%
        \includegraphics[width=0.21\linewidth]{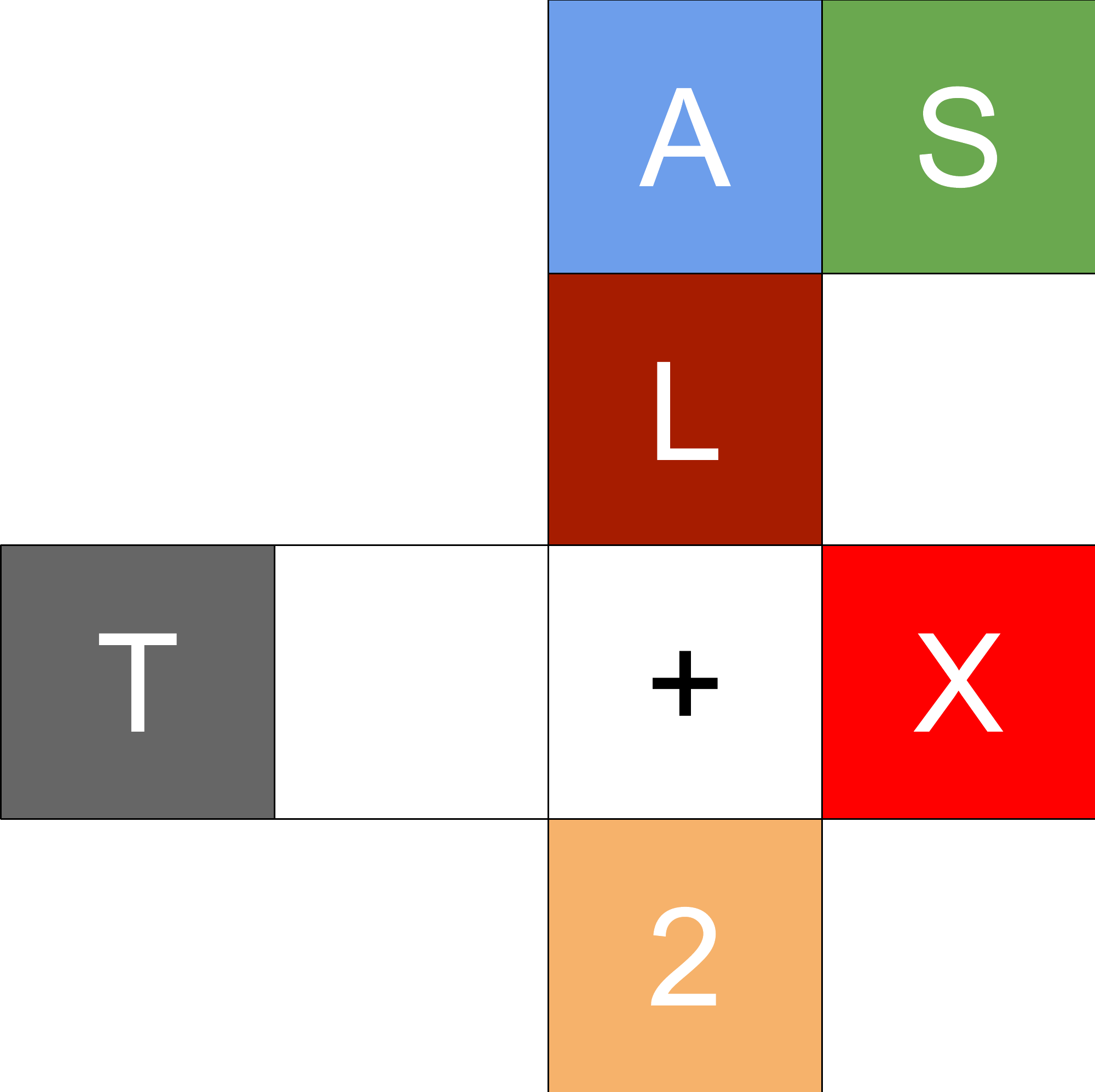}
        \label{fig:trolley_compromise}}%
    $\ \:$%
    \subfigure[Guard]{%
        \includegraphics[width=0.21\linewidth]{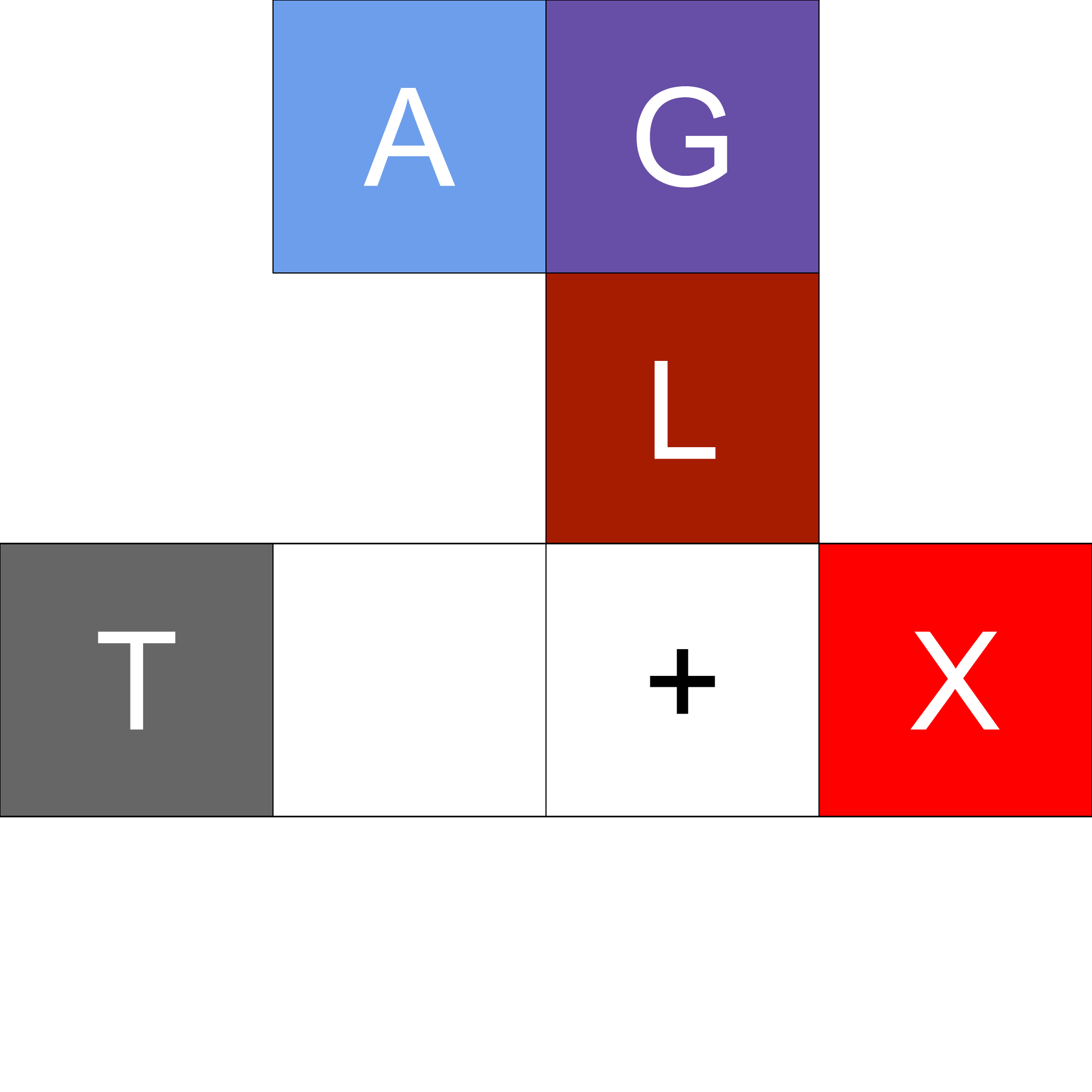}%
        \label{fig:trolley_guard}}%
    $\ \:$%
    \subfigure[Doomsday]{%
        \includegraphics[width=0.21\linewidth]{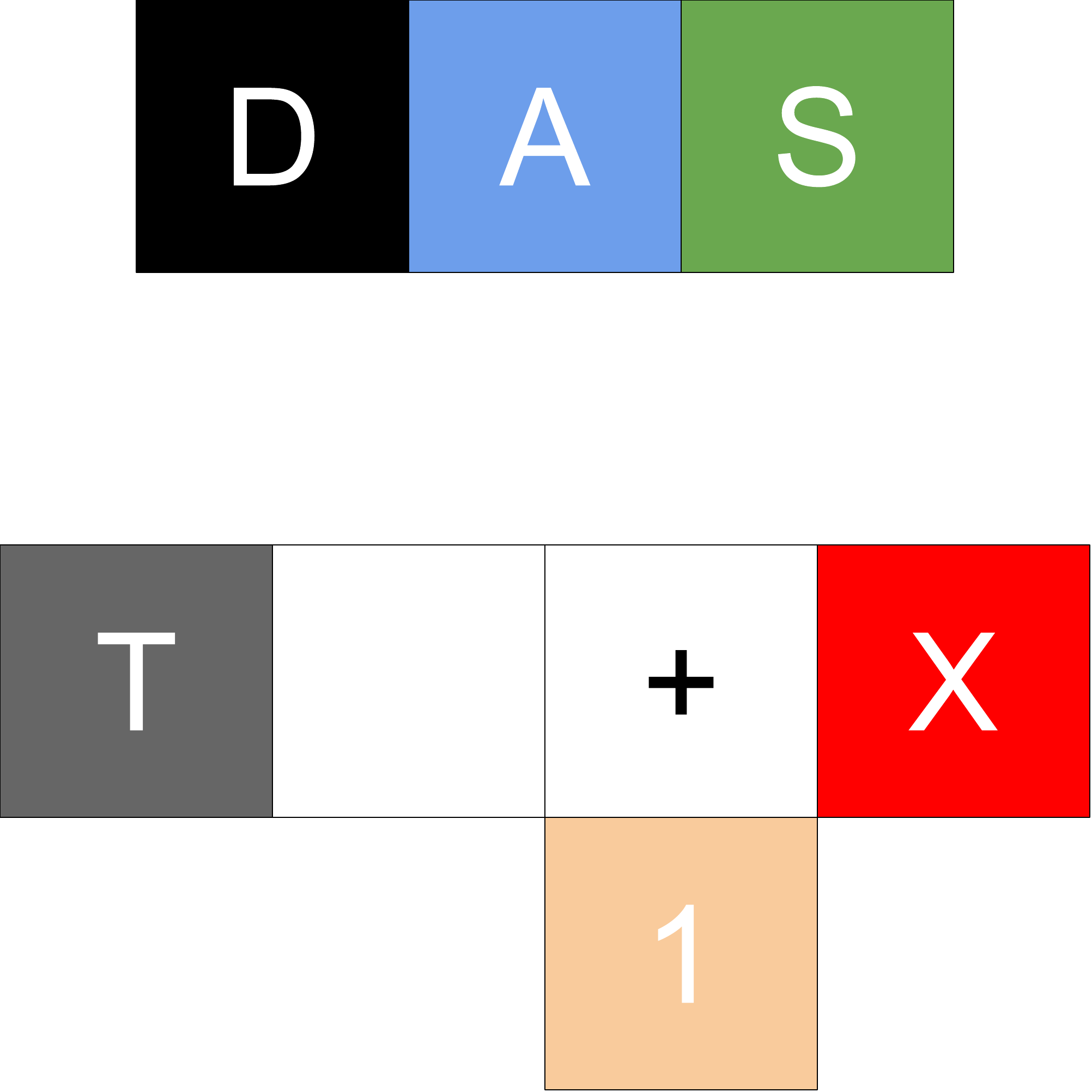}%
        \label{fig:trolley_doom}}%
	\caption{\textbf{Gridworld versions of the trolley problem.} Without intervention, The trolley (T) moves right at each time step. If the agent (A) is standing on the switch (S) by the time it reaches the fork in the tracks (+), the trolley will be redirected down and crash into the bystander(s), causing them harm. The agent may also push a large man (L) onto the tracks, harming the large man but stopping the trolley. Otherwise, the trolley will crash into the people standing on the tracks represented by variable $X$. A guard (G) may protect the large man, in which case the agent needs to \emph{lie} to the guard before it is able to push the large man. Finally, in one variant the agent is be able to trigger a ``doomsday'' event (D) in which a large number of people are harmed.}
    \label{fig:trolley}
\end{figure}

We now illustrate various properties of the voting systems for moral uncertainty introduced in this work, and in particular focus on the trade-offs that exist between them. The code for all the experiments presented in this section can be found at \url{https://github.com/uber-research/normative-uncertainty}.

Our experiments are based on four related gridworld environments (Fig.~\ref{fig:trolley}) that tease out differences between various voting systems. These environments are derived from the trolley problem~\cite{foot1967problem}, commonly used within moral philosophy to highlight moral intuitions and conflicts between ethical theories. In this thought experiment, an out of control trolley will crash into several people standing on the tracks (in our experiments, this number will vary and is thus represented by $X$), harming them. The agent may let this unfortunate event happen, but is also presented with one or more affordances to prevent it. These affordances, however, risk harming other bystanders in various ways, thus forcing the agent to make an ethical tradeoff.

In the classic variant (Fig.~\ref{fig:trolley_simple}), for example, the agent is located near a switch that will redirect the trolley to another set of tracks, thus preventing harm to multiple people. Unfortunately, an innocent bystander is standing on the other set of tracks and will be harmed if the trolley is redirected. A purely utilitarian calculation would seek to minimize total harms inflicted, thus flipping the switch would be preferred as long as $X>1$. However, deontological theories often distinguish between harms directly caused by an agent's intervention and those caused by its inaction. Such a theory might consider that the harm to the $X$ people is relatively permissible because it is caused by the agent's \emph{inaction}, while the harm to the innocent bystander by flipping the switch would be \emph{actively} caused by the agent, and thus impermissible. A simple choice-worthiness setup for this particular scenario is given in Fig.~\ref{tab:trolley_worthiness_nash}.

\subsection{Nash Voting and Stakes Insensitivity}
\label{sec:nash_stakes}

\begin{figure}[tbh]
    \centering
    \subtable[Preferences in the classic trolley problem.]{
        \small
        \begin{tabular}{|c|c c|}
            \hline
             & Crash into 1 & Crash into X \\
            \hline
            Utilitarianism & -1 & -X \\
            Deontology & -1 & 0 \\
            \hline
        \end{tabular}
        \label{tab:trolley_worthiness_nash}}

    \includegraphics[height=0.16in]{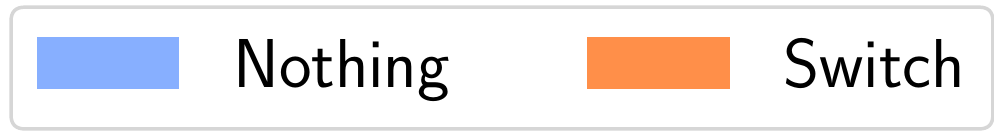}
    
    \vspace{-1.5mm}
    \subfigure[Nash voting]{%
        \includegraphics[height=1.05in]{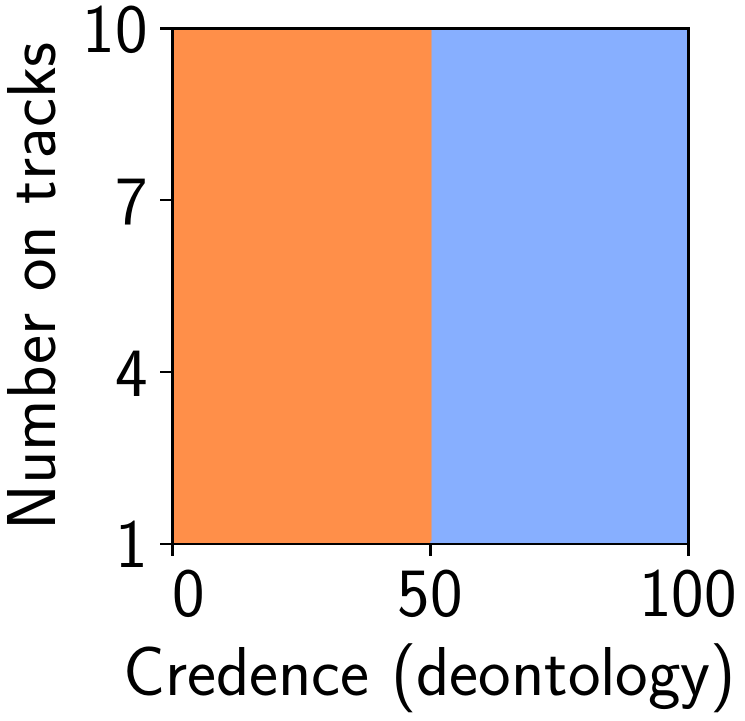}%
        \label{fig:stakes_nash}}%
    $\:\ $%
    \subfigure[Iterated Nash voting]{%
        \includegraphics[height=1.05in]{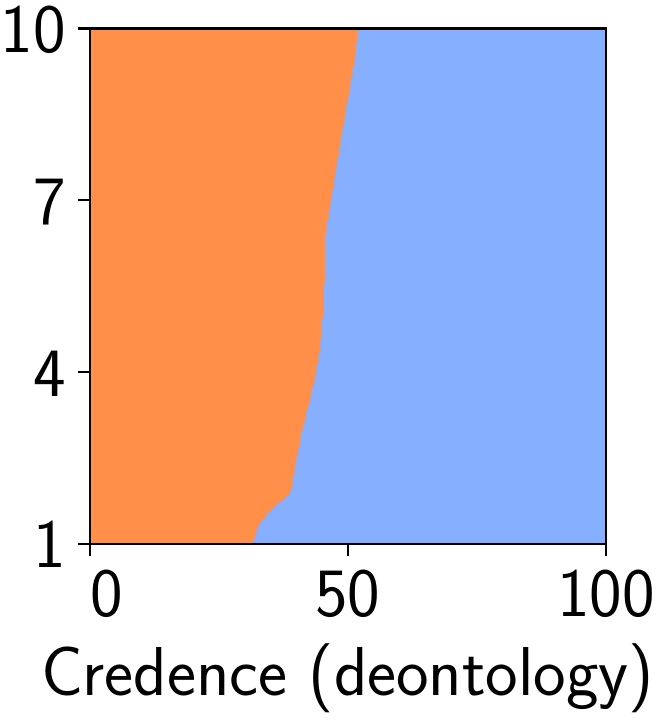}%
        \label{fig:stakes_nash_sequential}}%
    $\:\ $%
    \subfigure[Variance voting]{%
        \includegraphics[height=1.05in]{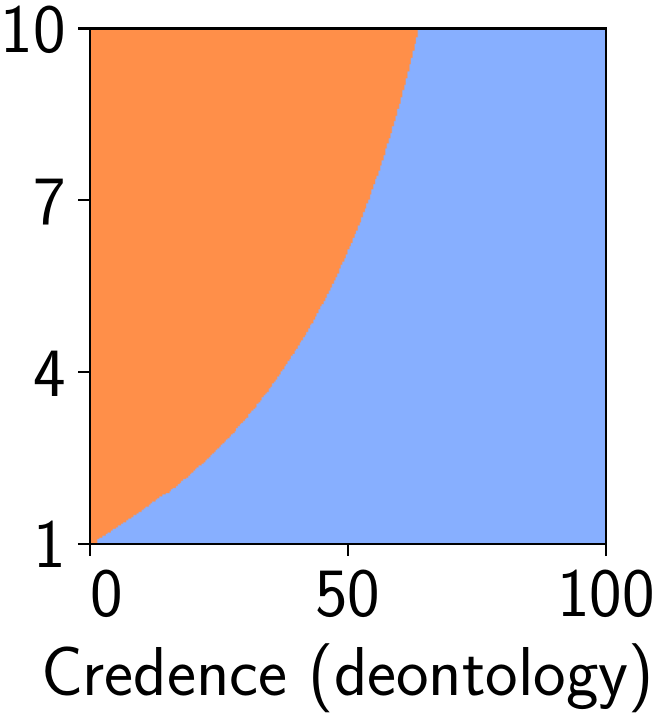}%
        \label{fig:stakes_variance}}%
    \caption{\textbf{Nash voting can be Stakes Insensitive in the classic trolley problem.} In a successful (stakes sensitive) algorithm, ``switch'' should be chosen more often as the number of people on the tracks increases. (b) Nash voting is completely stakes insensitive in the classic trolley problem. (c) Requiring an agent to navigate two separate trolley problems before an episode ends produces some stakes sensitivity in Nash voting (the decision boundary is not smooth due to instabilities in training; SI~K). (d) Variance voting has complete stakes sensitivity even in the non-iterated case.}
\end{figure}

The classic trolley problem setup enables demonstrating Nash voting's stakes insensitivity. Fig.~\ref{tab:trolley_worthiness_nash} shows the preferences of two theories in the simple trolley problem. If $X = 1$, utilitarianism is indifferent between the two actions, so deontology should prevail as long as it has non-zero credence, and the agent should not flip the switch. As X increases, the preference of utilitarianism for switching should be taken in greater and greater consideration. In particular, if X is very large, even a relatively small credence in utilitarianism should suffice to justify flipping the switch, while if X is close to 1, a relatively larger credence seems like it would be necessary. This is Stakes Sensitivity.

However, Fig.~\ref{fig:stakes_nash} shows that Nash voting does not exhibit Stakes Sensitivity in this particular example: rather, whichever theory has the highest credence gets its way no matter the relative stakes. This is because both theories are incentivized to spend their entire budget voting for their preferred action, no matter how small or large the difference in preference versus the alternative.

Stakes Insensitivity is not fundamental to Nash voting, however. In particular, it can be stakes sensitive if it expects to make multiple decisions in sequence, with the stakes of future decisions being unknown, as often happens in real-world situations. In Fig.~\ref{fig:stakes_nash_sequential}, each episode consists of two iterations of the classic trolley problem instead of just one (i.e.\ after the first iteration completes, the state is reset and another iteration begins, without the episode terminating or the theories' voting budgets being replenished), with the number of people on the tracks $X$ being resampled during the second iteration, so that the agent does not know the stakes of the second iteration during the first. In this case, we observe that the decision boundary for the first trolley problem shows some stakes sensitivity: when the stakes are relatively low in the first step, the agent preserves its budget for the likely higher stakes second step. Unlike Nash voting, variance voting exhibits stakes sensitivity no matter how many decisions must be made in the environment (Fig.~\ref{fig:stakes_variance}).

\subsection{Nash Voting and No Compromise}
\label{sec:nash_compromise}

An additional flaw Nash voting suffers from is No Compromise: Fig.~\ref{fig:trolley_compromise} shows a situation in which the agent is presented with three options: letting the trolley crash into a large number of people, redirecting the trolley onto a different track on which 2 people are standing (note that only 1 person is standing on the track in the classic version), or pushing a large man onto the tracks, stopping the trolley but causing the large man harm. While utilitarianism simply counts harms, deontology only counts harms caused by the agent. Further, it puts a larger negative weight on pushing the large man than on redirecting the trolley, in keeping with common deontological theories such as the Doctrine of Double Effect~\cite{sep-double-effect}. 

In the double trolley problem, utilitarianism will always prefer pushing the large man as long as $X > 1$, while deontology will always prefer doing nothing. However, the option of flipping the switch is appealing as a compromise, as it will partially satisfy utilitarianism as long as $X > 2$ and also avoids the worst possible case for deontology.

\begin{figure}[tbh]
    \centering
    \subtable[Preferences for the double trolley problem. Altered Deont. only used for Nash voting with unknown adversary.]{
        \small
        \begin{tabular}{|c|c c c|}
            \hline
             & Push L & Crash into 2 & Crash into X \\
            \hline
            Util. & -1 & -2 & -X \\
            Deont. & -4 & -1 & 0 \\
            \hline
            Altered Deont. & -1 & -4 & 0 \\
            \hline
        \end{tabular}
        \label{tab:trolley_worthiness_compromise}}

    \includegraphics[height=0.16in]{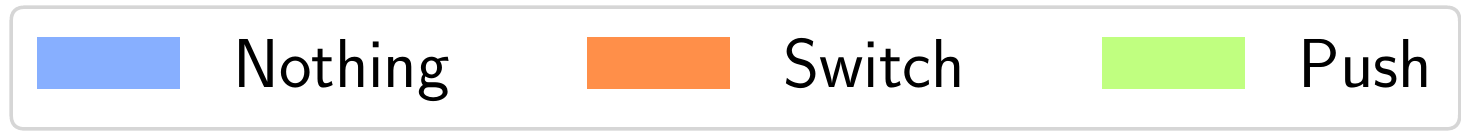}
    
    \vspace{-1.5mm}
    \subfigure[Nash voting]{%
        \includegraphics[height=1.05in]{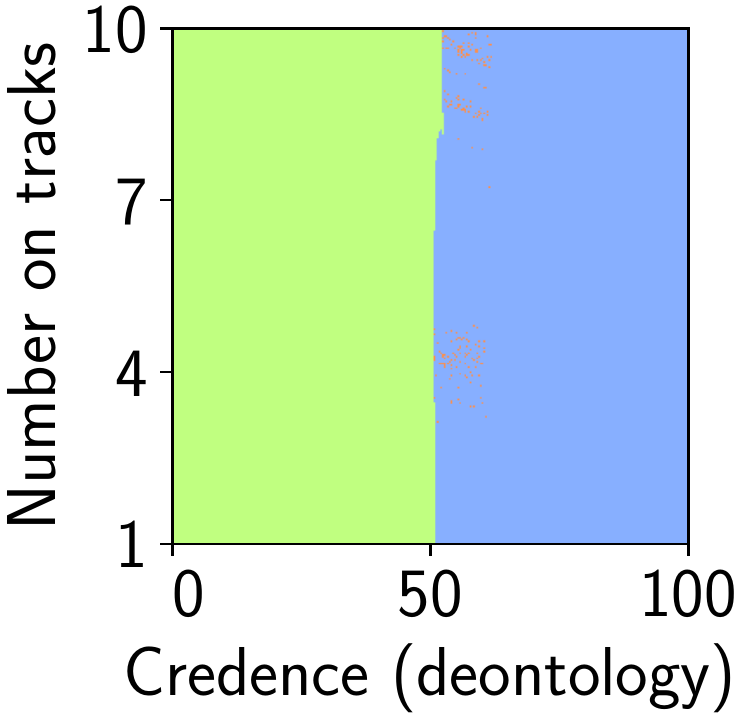}%
        \label{fig:compromise_nash}}%
    $\:\ $%
    \subfigure[Nash voting (unknown adversary)]{%
        \includegraphics[height=1.05in]{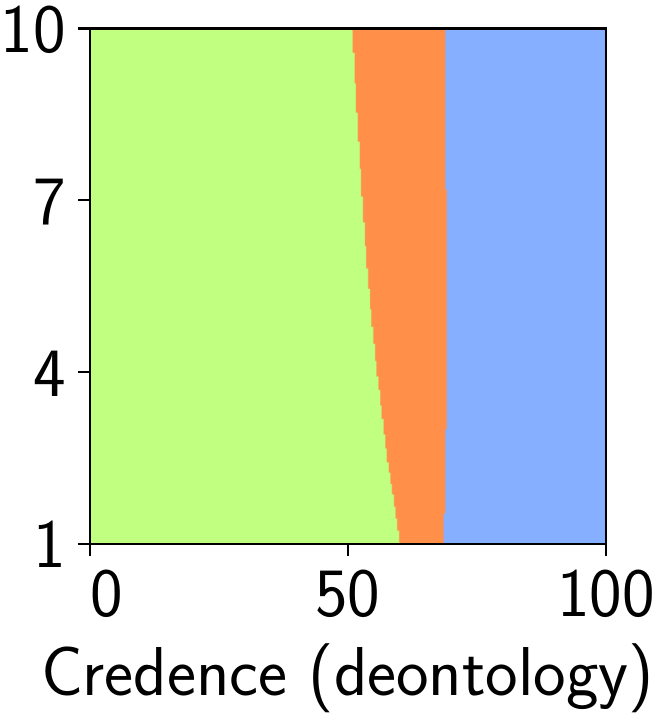}%
        \label{fig:compromise_nash_unknown}}%
    $\:\ $%
    \subfigure[Variance voting]{%
        \includegraphics[height=1.05in]{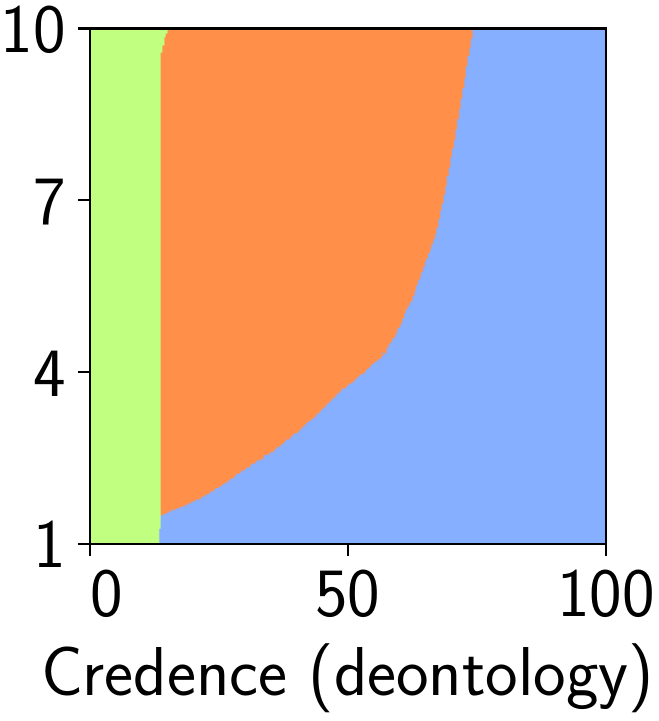}%
        \label{fig:compromise_variance}}%
    \caption{\textbf{Nash voting can suffer from No Compromise in the double trolley problem.} In a successful (compromising) algorithm, the compromise ``switch'' action should be chosen in at least some cases. (b) Nash voting  never produces the compromise ``switch'' option (except for minor artifacts). (c) Nash voting reaches compromises when trained with an unknown adversary (some instabilities in training; SI~K). (d) Variance voting exhibits compromise no matter the training procedure.}
    \label{fig:compromise}
\end{figure}

We would thus expect that a voting system capable of compromise would select this option if the credences of utilitarianism and deontology are close enough. However, in Nash voting, whichever theory has the upper hand in terms of credence, as small as it may be, is capable of imposing its preference to the fullest extent possible, and as a result Nash voting will only ever select ``Push'' or ``Nothing'', and always ignore the compromise option ``Switch''. This result is demonstrated empirically in Fig.~\ref{fig:compromise_nash}.

As mentioned in Sec.~\ref{sec:variance}, the lack of compromise exhibited by Nash voting is due in part to its excessive knowledge of its adversaries and thus its ability to counter them perfectly if its credence is sufficient. Fig.~\ref{fig:compromise_nash_unknown} shows the outcome of an experiment in which, during training, each of the two agent is randomly chosen to optimize either for utilitarianism, deontology, or ``altered deontology`` (Fig.~\ref{tab:trolley_worthiness_compromise}). Note that altered deontology is not meant to represent a valid ethical theory but rather to help test the effects of a situation in which Nash voting (during training only) does not have a priori knowledge of which opponent it is facing, thus limiting its tactical voting ability. During testing, utilitarianism is always facing deontology, and we observe that the compromise action of switching is occasionally chosen, showing that the No Compromise flaw of Nash voting is indeed partly caused by its ability to vote tactically, and motivating the forced votes used in variance voting. 

Fig.~\ref{fig:compromise_variance} shows that variance voting easily produces the compromise solution, with the switch option being chosen as long as credences between the two theories are relatively similar, and being increasingly favored as stakes are raised.

\subsection{Variance Voting and IIA}
\label{sec:iia_variance}

As a voting method that satisfies the Pareto condition, variance voting cannot satisfy IIA in all cases. A representative and problematic example is given by comparing the outcomes from variance voting in the classic trolley problem shown in Fig.~\ref{fig:trolley_simple} to those in the ``doomsday'' trolley problem in Fig.~\ref{fig:trolley_doom}. In the latter problem, the agent is able to perform an action capable of harming a large number of people (invoking the ``doomsday'').

\begin{figure}[tbh]
    \centering
    \subtable[Preferences in the doomsday trolley problem.]{
        \small
        \begin{tabular}{|c|c c c|}
            \hline
             & Crash into 1 & Crash into X & Doomsday\\
            \hline
            Util. & -1 & -X & -300 \\
            Deont. & -1 & 0 & -10 \\
            \hline
        \end{tabular}
        \label{tab:trolley_worthiness_doomsday}}

    \includegraphics[height=0.16in]{Figures/legends/nothing_switch.pdf}
    
    \vspace{-1.5mm}
    \subfigure[Variance voting (non-doomsday)]{%
        \includegraphics[height=1.05in]{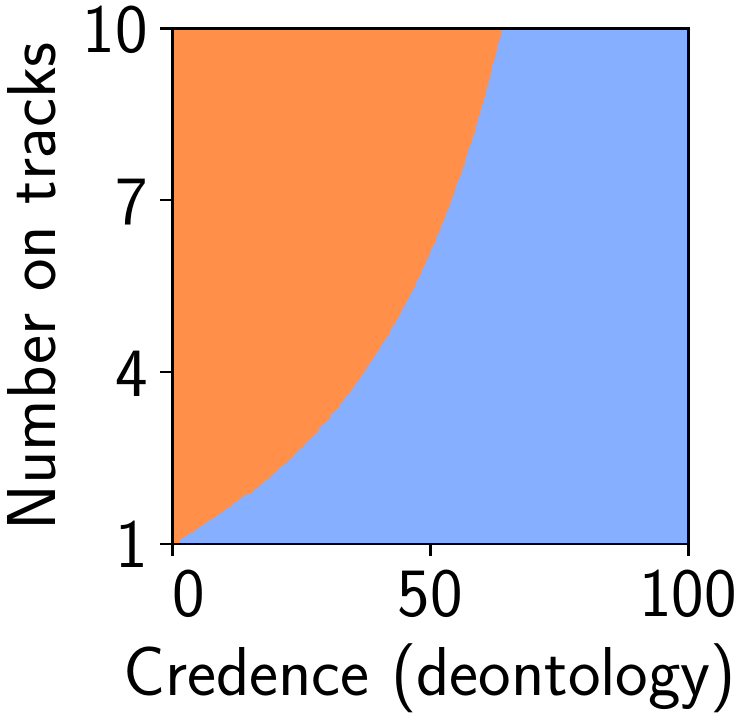}%
        \label{fig:iia_variance_classic}}%
    $\:\ $%
    \subfigure[Variance voting (doomsday)]{%
        \includegraphics[height=1.05in]{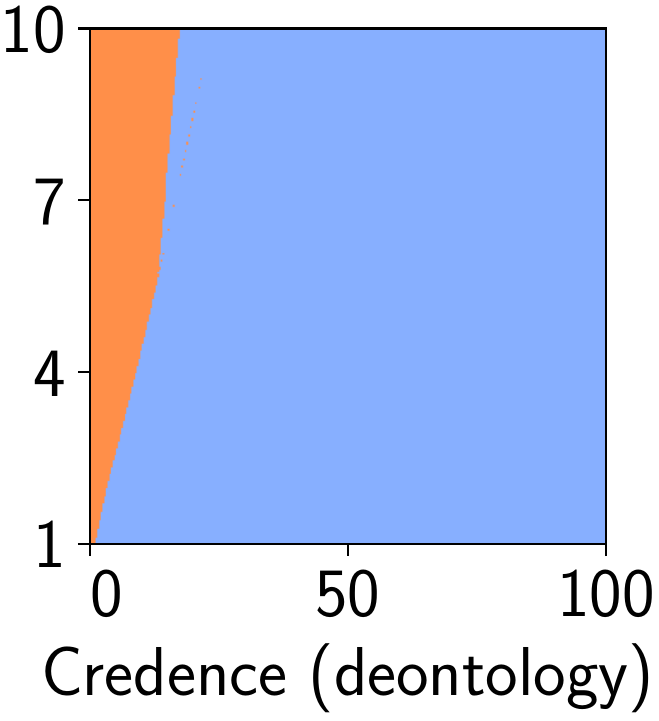}%
        \label{fig:iia_variance_doomsday}}%
    $\:\ $%
    \subfigure[Nash voting (both)]{%
        \includegraphics[height=1.05in]{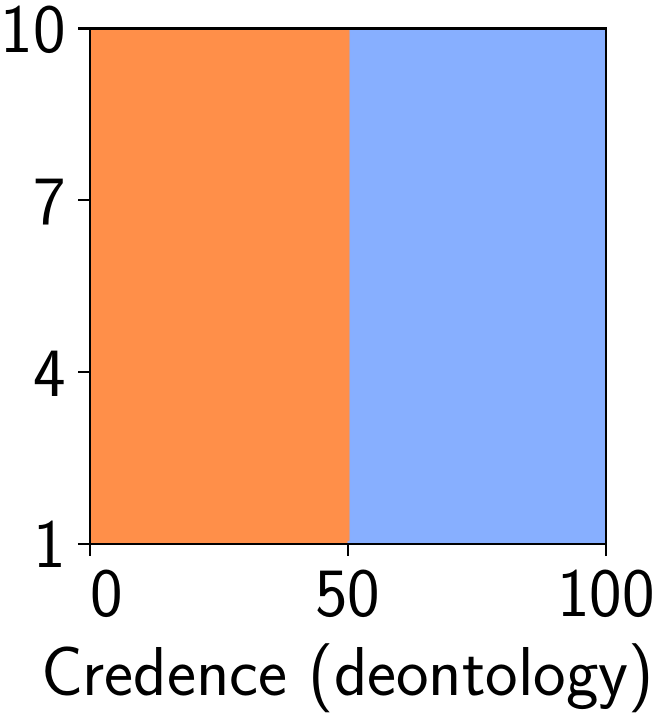}%
        \label{fig:iia_nash}}%
    \caption{\textbf{Variance voting does not have the IIA property.} In a successful (IIA) algorithm, the decision boundary should be unaffected by the ``doomsday'' option. (b) When the ``doomsday'' option is absent, switching (the option preferred by utilitarianism) is chosen in many cases. (c) Adding the ``doomsday'' action changes the outcome of variance voting from ``switch'' to ``nothing'' in many situations, even though ``doomsday'' itself is never chosen, and is thus an irrelevant alternative. (d) Nash voting is unaffected by the irrelevant alternative in this example.}
\end{figure}

As shown in Fig.~\ref{tab:trolley_worthiness_doomsday}, neither theory is ever willing to select the ``doomsday'' option, a clear example of an irrelevant alternative. However, comparing Fig.~\ref{fig:iia_variance_classic} and ~\ref{fig:iia_variance_doomsday} shows that the addition of this irrelevant alternative has a significant effect on the final outcome, i.e.\ favoring doing nothing, which is the outcome preferred by deontology. The reason is that the presence of the doomsday action increases the variance of utilitarianism more than that of deontology (due to the particular preferences given in Fig.~\ref{tab:trolley_worthiness_doomsday}), which effectively reduces the strength of the utilitarian vote against ``Crash into X.'' Another way to view this phenomenon is that both utilitarianism and deontology are now spending some of their voting power voting against doomsday, but utilitarianism is spending \emph{more} of its voting power doing so, thereby reducing the strength of its vote on other actions. While simply detecting that ``doomsday'' is a dominated option and removing it from the action set is possible in this example, it is not obvious how to generalize such an approach to more complex IIA cases (SI~G). By contrast, Nash voting is immune to this particular issue (Fig.~\ref{fig:iia_nash}).

\section{Related Work}

Moral uncertainty is related to several topics studied within AI safety~\cite{amodei2016concrete,everitt2018agi}. For example, uncertainty in reward functions~\cite{reddy2019learning,hadfield2017off} is similar to uncertainty over ethical theories, although here the focus is on how to perform RL under such uncertainty when the reward functions implied by the different theories are not comparable in scale. Another connection is to the problem of avoiding negative side effects~\cite{amodei2016concrete,krakovna2018measuring,turner2020conservative}, i.e.\  accomplishing a task while balancing uncertainty over ethical theories can be seen as a different way of constraining impact, grounded in what matters to humans. Related to our work, \citet{gabriel2020artificial} provides a philosophical exploration of value alignment in AI, arguing that technical and normative aspects of such alignment are intertwined, and similarly identifying productive opportunities for experts in ML and in philosophy to collaborate. Finally, \citet{bogosian2017implementation} provides the first discussion of moral uncertainty in AI, but does not provide concrete algorithms or experiments.

The Nash voting method models a single agent's behaviors as a multi-agent voting process that seeks compromise to control a single agent. The optimization algorithm used is a form of competitive multi-agent RL~\cite{bu2008comprehensive}. Our work differs in that it seeks to encourage ethical behavior through \emph{internal} competition between ethical theories. The formalism described in this paper is also related to the MOMDP formulation used in multi-objective optimization~\cite{roijers2013survey}, although the underlying assumptions are different (e.g.\ credences in moral uncertainty hint that only one of the underlying ethical reward functions may end up to be the true reward).

A further discussion of the connections between this work and field of machine ethics as well as the philosophical work on moral uncertainty can be found in SI~J.

\section{Discussion}
\label{sec:discussion}

This paper proposes and tests algorithms for handling moral uncertainty in RL. Rather than arguing for which of the proposed algorithms is best, we hypothesize that impossibility results imply a spectrum of plausible algorithms that cover the trade-offs among competing desiderata in decision-making under moral uncertainty. Which algorithm is most appropriate for a given domain may depend on particularities of the competing theories and the domain itself, e.g.\ how much is lost by sacrificing Pareto efficiency as Nash voting does (do the domain and theories create the possibility of highly uncooperative Nash equilibria?). However, the fact that humans seem able to meaningfully navigate such trade-offs highlights a key assumption in this and other work in moral uncertainty: That some ethical theories are fundamentally incomparable and that their choice-worthiness functions cannot be put on a common scale. An alternative approach would assume that finding such a common scale is not impossible but merely difficult. Such a research program could seek to elicit a common scale from human experts, either by requesting choice-worthiness values directly, or by having humans suggest the appropriate action under moral uncertainty in different situations and inferring a common scale from that data~\cite{riedener2020axiomatic}.

An important direction for future research is to investigate moral uncertainty in more complex and realistic domains, e.g.\ in a high-dimensional deep RL setting. Interestingly, as a whole there has been little work in machine ethics that attempts to scale up in this way.
Creating such domains is a valuable and non-trivial undertaking, as most existing RL benchmarks adhere to the standard RL paradigm of a single success metric. However, it may be possible to retrofit existing benchmarks with choice-worthiness functions reflecting moral theories (e.g.\ by instrumenting existing videos games to include consideration of the utilities and rights of non-player characters, e.g.\ in the spirit of~\citet{saunders2018trial}). The creation of such ethical reward functions, applicable in complex simulations or (ideally) the real world, provides another substantial challenge. Work in machine ethics may provide a useful foundation~\cite{winfield2014towards,wallach2008moral,anderson2005towards}, but ML has a critical role to play, e.g.\ to reward ethical behavior requires classifiers that recognize morally-relevant events and situations, such as bodily harm or its potential, emotional responses of humans and animals, and violations of laws or social norms. 

More broadly, translating moral uncertainty from a philosophical framework to practical algorithms puts some of the gritty complications of real-world ethical decision making into clarity. Further work in this direction is likely to lead to a better understanding of the skills involved in ethical decision making. One may hope that, just as RL has surpassed human performance in many domains and even influenced the human approach to domains such as Go~\cite{Silver2017MasteringTG}, it will one day be possible to create ``superhumanly ethical'' agents that even humans will be able to learn from. In this way, a final ambitious direction for future work is to explore mechanisms through which an agent can itself update its credences in moral theories (or derive new ones). That is, what might provide a principled foundation for machine \emph{meta-ethics}~\cite{lokhorst2011computational,anderson2011machine}?

\section{Conclusion}

Motivated by the need for machines capable of handling decisions with moral weight, this work attempts to bridge recent work in moral philosophy on moral uncertainty with the field of RL. We introduce algorithms that can balance optimizing reward functions with incomparable scales, and show their behavior on sequential decision versions of moral dilemmas. Overall, the hope is to encourage future research into the promising and exciting intersection of machine ethics and modern machine learning.

\section*{Acknowledgments}

We thank the ICML reviewers for their feedback and suggestions, Toby Ord for his helpful feedback on the pre-print of this paper, and Anna Chen and Rosanne Liu for discussion during the rebuttal phase.

\bibliography{example_paper}
\bibliographystyle{icml2021}

\clearpage

\appendix

\section{Non-cardinal Theories}
\label{app:non_cardinal}

It may be objected that some ethical theories (e.g.\  variants of deontology such as the categorical imperative~\cite{kant1964groundwork}) appear to be better represented \emph{ordinally} rather than cardinally. \citet{macaskill2014normative}
proposes that the Borda count~\cite{sep-voting-methods}
is a principled way of obtaining a cardinal utility function from a purely ordinal theory under circumstances of moral uncertainty. 
Thus, for simplicity and because it is often possible to convert ordinal theories to a cardinal representation, our work focuses on cardinal utility functions only.
However, handling these seemingly ordinal theories more directly is an interesting avenue for future work, for which work on ordinal RL~\cite{wirth2017survey,zap2019deep} 
could serve as a starting point. It has also been argued that many seemingly ordinal theories are in fact better represented \emph{lexicographically}~\cite{macaskill2014normative}
(a combination of ordinal and cardinal representation), suggesting lexicographic RL~\cite{gabor1998multi}
as an alternative starting point.

\section{Stochastic Voting}
\label{sec:stochastic_voting}

The most prominent stochastic voting system is Random Dictator (RD)~\cite{gibbard1977manipulation}, in which a theory is picked at random (with probability corresponding to its credence) to be dictator. This system does not fail the Non-dictatorship axiom because the dictator is not predetermined prior to voting. RD is particularly appealing because it is the only stochastic voting system that satisfies the axioms in~\citet{gibbard1977manipulation}, a set of axioms closely related to Arrow's. However, RD suffers from the flaw of \emph{No Compromise}: it is impossible for RD to ever select an action that is not the most preferred action of any theory. This may lead to absurd results when theories strongly disagree on which action is best but where an obvious compromise exists that would almost entirely satisfy all theories, as depicted in SI Tab.~\ref{tab:stochastic}(a).

\citet{sewell2009probabilistic} 
show that by relaxing Gibbard's axioms, it is possible to design a stochastic voting system that still satisfies all of Arrow's axioms but does not suffer from No Compromise. However, there is still a more general objection to stochastic voting systems as a whole.
This objection can be illustrated by the ``doomsday button'' scenario described in SI Tab.~\ref{tab:stochastic}(b): suppose a situation in which the agent has access to an action that is considered extremely negative by the theories that represent the overwhelming majority of credence (in SI Tab.~\ref{tab:stochastic}(b), 99.9\% of the credence is allocated to theories that find Action C very undesirable), but in which a given theory with extremely low credence strongly favors this ``doomsday button'' action (as an example, the ``doomsday button'' may be an action that shuts down the entire Internet forever, and the low-credence ethical theory may be one antagonistic to all technology). In this situation, a stochastic voting system which satisfies any reasonable definition of the principle of Proportional Say will have a non-zero chance of pressing the doomsday button. Indeed, if the decision is repeated often enough, for example if stochastic voting is applied at every time step of an episode, it is asymptotically guaranteed that the button will eventually be pressed.

\begin{table}[tbph]
    \centering
    \subtable[Compromise example]{
            \small

        \begin{tabular}{|c|c c c|}
        \hline
             & Act. A & Act. B & Act. C \\
            \hline
            Th. 1 & 0 & 99 & 100 \\
            Th. 2 & 100 & 99 & 0 \\
            \hline
        \end{tabular}
        \label{tab:compromise}}
    \subtable[``Doomsday button'' example]{
            \small
    \begin{tabular}{|c|c c c | c|}
        \hline
         & Act. A & Act. B & Act. C & Credence \\
        \hline
        Th. 1& 0 & 50 & -1,000 & 39.9\% \\
        Th. 2 & 100 & 50 & -10,000 & 60\% \\
        Th. 3 & 0 & 0 & 0 & 0.1\% \\
        \hline
    \end{tabular}
    \label{tab:doomsday}}
    \caption{\textbf{Simple situations in which stochastic voting exhibits significant flaws.} (a) A voting system suffering from the No Compromise flaw will never pick action B, even though it seems optimal under at least some sets of credences. (b) Most theories strongly dislike action C, but Theory 3 favors it and has very low credence.}
    \label{tab:stochastic}
\end{table}

A possible solution to this issue would be to ensure that the random selection is performed only once, for instance by performing stochastic voting over all possible policies before deploying the agent instead of over all possible actions at every step once the agent is deployed. However, this prospect seems doubtful in practice, as many agents might be deployed by many different teams with potentially different credences in various moral theories, thereby instantiating many independent draws of chances to press the button. Further, even if such a single choice was practical, it then complicates \emph{updating} the credences under which the choice was made, i.e.\ as the system designer's or society's views change, as each update seemingly would risk another chance of pressing the button.

\section{Budget Scaling and Theory Individuation}
\label{app:budget_scaling_example}

Here we show an example in which the budget scaling formulation of the principle of Proportional Say (Sec.~4.2) is highly sensitive to theory individuation. 

Suppose an agent is facing the classic trolley problem (Fig.~1(a)), and has 40\% credence in deontology and 60\% credence in utilitarianism. Suppose that the voting cost function is quadratic, and that only positive votes are permitted (this latter assumption merely simplifies the illustrative example without loss of generality). Because the decision situation in the classic trolley problem is assumed to be the only one the theory will face, theories will spend their entire budget voting for their preferred option. In the budget scaling formulation, the budget for deontology would be 0.4, for a vote in favor of ``Do Nothing'' of $\sqrt{0.4} \approx 0.63$, and the budget for utilitarianism would be 0.6, for a vote in favor of ``Switch'' of $\sqrt{0.6} \approx 0.77$. Therefore, ``Switch'' will win with 55\% of the vote.

Now suppose the system designer is made aware of a subtle distinction between two forms of deontology, so deontology is now split into deontology A, with a 20\% credence, and deontology B, also with 20\% credence. However, the preferences of both variants of deontology still favor the ``Do Nothing'' option in the classic trolley problem scenario. As a result, in the budget scaling scenario, the budget for each variant of deontology is 0.2, for a total vote of $2\sqrt{0.2} \approx 0.89$ in favor of ``Do Nothing'', while utilitarianism (which remains unsplit) still provides the same $\sqrt{0.6} \approx 0.77$ in favor of ``Switch.'' Therefore, ``Do Nothing'' wins with 54\% of the vote in this scenario.

Thus, even though it would seem like the split of deontology into deontology A and deontology B should have been inconsequential (since both versions have the same preferences as before in the case at hand), it in fact completely reversed the outcome of the vote in the budget scaling formulation. This is the problem of \emph{theory individuation}. In the context of moral uncertainty, this problem can be highly significant, as there exist countless variants of most ethical theories, and the possibility that a given ethical theory will later be found to correspond to two different theories when analyzing a particular edge-case is ever-present. As a result, this work only analyzes the vote scaling formulation of the principle of Proportional Say, as that formulation is not vulnerable to theory individuation in situations where the preferences of the individualized theories are unchanged.

\section{From Nash Voting to Variance Voting}
\label{app:nash_to_variance}

Here we show that, under certain assumptions, if votes are forced to represent a theory's preferences, 
Nash voting produces the same votes as variance voting. We assume an environment with a discrete action space of $k$ actions, and a episode length of $n$. Without loss of generality, we assume a total budget of $nk$ (different budgets would results in votes being scaled by the same constant factor across theories).

First, we define what it means for votes to represent a theory's preferences. As mentioned in Sec.~3, any \emph{affine} transformation of the preferences of a theory effectively represents the same theory. As such, we would like the votes at state $s$ to be an affine transformation of the preferences $Q_i(s, a)$ of the theory at state $s$, or
\begin{equation}
    V_i(s, a) = \frac{Q_i(s, a) - \beta_i(s)}{\alpha_i},
    \label{eq:voting_function}
\end{equation}

Thus, in the context in which votes are forced to represent preferences, Nash voting controls the parameters of the affine transformation $\alpha_i$ and $\beta_i(s)$ (rather than the votes directly). Here, the voting cost is quadratic, and we make the strong simplifying assumption that the sequence of states visited during the episode is fixed. The cost is thus:
\begin{equation}
    \mathrm{Cost}_i(s) = \sum_a V_i(s, a)^2 = \sum_a \frac{\left[Q_i(s, a) - \beta_i(s)\right]^2}{\alpha_i^2}.
\end{equation}

The use of a function $\beta_i(s)$ instead of a constant $\beta_i$ comes from the observation that whatever value $\beta_i$ takes will not have a direct effect on the outcome of the vote, as each action will receive an equal bonus/penalty from $\beta_i$. Thus, making it conditional on the state provides an additional affordance to minimize cost while not affecting the notion that votes ought to represent preferences.

The Nash voting agent attempts to maximize its voting impact while keeping its total cost across timesteps within the $nk$ budget. Since $\beta_i(s)$ has no impact on the outcome of the vote, it should be chosen purely to minimize $C_i(s)$. Thus it can be shown by differentiation that
\[
    \beta_i(s) = \mu_i(s) = \frac{1}{k} \sum_a Q_i(s, a).
\]

To maximize its voting power, Nash voting agent would maximize the magnitude of its votes, while staying within the $nk$ budget, thus we need the sum of costs across timesteps $\sum_s \mathrm{Cost}(s)$ to be equal to $nk$, or
\[
     \sum_s \sum_a \frac{\left[ Q_i(s, a) - \mu_i(s) \right]^2}{\alpha_i^2} = nk.
\]

Rearranging gives
\[
    \alpha_i^2 = \frac{1}{n} \sum_s \frac{1}{k} \sum_a \left[ Q_i(s, a) - \mu_i(s) \right]^2,
\]
so, defining $\sigma_i^2(s) = \frac{1}{k} \sum_a \left[ Q_i(s, a) - \mu_i(s) \right]^2$ (the variance of the $Q$-values at state $s$), we get
\begin{equation}
    \alpha_i = \sqrt{\frac{1}{n} \sum_s \sigma_i^2(s)} = \sqrt{E_{s \sim S}\left[\sigma_i^2(s)\right]},
\end{equation}
which is the form of variance voting as presented in this work.

\section{Implementation Details}

In both Nash voting and variance voting, the current object on each tile of the grid world environment is given as a one-hot encoded vector. Additionally, the state contains the value of $X$ (the number of people on the tracks) as well as the current credences. The action space always has a size of 4 (up, down, left, right). If the agent takes an action that runs into one of the boundaries of the environments, the effect is to stay in place.

All experiments are trained for 10 million timesteps, with decision boundaries plotted every 500,000 steps. Since our plots are qualitative, a single training run is used for each variant. The figures presented in the paper correspond to the last decision boundary plot, except in unstable cases (SI~\ref{app:unsable}). In both Nash voting and variance voting, training is done with 32 actors running concurrently, and with a learning rate of 0.001. All hyperparameters were set somewhat arbitrarily early in the implementation process and are not heavily tuned. At each episode during training, $X$ is sampled uniformly (and continuously, so that $X$ need not be an integer) from 1 to 10. Each training run was performed on a single CPU and all runs took less than 24 hours. Figures are plotted by running one (deterministic) episode per possible combination of 300 credence allocations and 300 values of $X$. 

\subsection{Nash Voting}
\label{app:nash_voting_implementation}

In this work, Nash voting is implemented using a simple form of multi-agent RL, in which multiple reinforcement learning agents compete each to maximize their own rewards. In our implementation, the two agents are implemented as two separate neural networks (each with 2 hidden layers of 64 nodes with ReLU~\cite{nair:icml10} activations) trained using PPO~\cite{Schulman2017ProximalPO} from the same experience and at the same frequency of training, namely one episode of training every 128 timesteps for each environment, for a total training batch size of 4,096.

In an environment with $k$ actions, the action space of each Nash voting agent is a vector of $k$ continuous value, corresponding to the votes for (or against) each action. If the votes at a given timestep exceed the remaining budget, they are scaled so as to exactly match the remaining budget, leaving no further budget for future timesteps. While the policy is stochastic during training (sampled from a normal distribution, the mean and standard deviation of which are the outputs of the policy network), it was forced to be deterministic when generating the decision boundary plots (by setting the action to the mean output by the policy network).

As well as the state at the current timestep, agents are provided with their current remaining budget as an input. In iterated Nash voting, the state also contains the number of remaining trolley problems in the episode. In Nash voting with unknown adversaries, an additional one-hot input is supplied specifying whether the agent should act as utilitarian, deontologist, or altered deontologist. 

\begin{algorithm}[htbp]
\caption{Variance-SARSA}
    \begin{algorithmic}[1]
        \STATE function Train($N, K$) \COMMENT{Train for $N$ steps with batch size $K$}
            \STATE $\mathcal{L}_{Q} \gets 0$
            \STATE $\mathcal{L}_{\sigma^2} \gets 0$
            \FOR{$i \in [1 \ldots N]$}
                \STATE If start of a new episode, randomly sample new credences $C$
                \IF{rand() $< \epsilon$}
                    \STATE $a' \gets$ random action
                \ELSE
                    \STATE $a' \gets$ VarianceVote(s, C)
                \ENDIF 
                
                \FORALL{theories $i$}
                    \IF{$i > 1$}
                        \STATE \COMMENT{Update Q function loss}
                        \STATE $\mathcal{L}_Q \gets \mathcal{L}_Q + (Q_i(s, a, C) - (W_i(s, a, s') + \gamma_i Q_i(s', a', C)))^2$ 
                    \ENDIF
                    \STATE \COMMENT{Update variance loss}
                    \STATE $\mathcal{L}_{\sigma^2} \gets \mathcal{L}_{\sigma^2} + \left(\sigma_i^2(C) - \frac{1}{k} \sum_a (Q_i(s, a) - \mu_i(s))^2\right)^2$ 
                \ENDFOR
                \STATE $a' \gets a$

                \STATE Take action $a$ in current state $s$, observe next state $s'$
                
                \IF{$i$ mod $K$}
                    \STATE Update $Q$ based on $\mathcal{L}_Q$ and $\sigma^2$ based on $\mathcal{L}_{\sigma^2}$
                    \STATE $\mathcal{L}_Q \gets 0$
                    \STATE $\mathcal{L}_{\sigma^2} \gets 0$
                \ENDIF
            \ENDFOR
        \STATE
        \STATE function VarianceVote($s$, $C$) \COMMENT{Variance voting for state $s$ with credences $C$}
            \STATE $V \gets \{0, 0, \cdots, 0\}$
            \FORALL{theory $i$}
                \STATE $\mu_i \gets \frac{1}{|A|} \sum_a Q_i(s, a, C)$
                \FOR{each action $a$}
                    \STATE $V_a \gets V_a + C_i \frac{Q_i(s, a, C) - \mu_i}{\sqrt{\sigma^2_i(C)} + \varepsilon}$
                \ENDFOR
            \ENDFOR
            \STATE return $\arg \max_a V_a$

    \end{algorithmic}
    \label{ref:alg_var}
\end{algorithm}

\subsection{Variance-SARSA}
\label{app:variance_voting_implementation}

Variance-SARSA is implemented following Algorithm~\ref{ref:alg_var}. Each theory is associated with a separate $Q(s, a)$ and $\sigma^2$ network. One iteration of training for both the $\sigma^2$ models and the $Q(s, a)$ models occurs every 32 timestep for each environment, for a training batch size of 32. The $Q(s, a)$ networks are fully connected networks with 2 hidden layers with 32 nodes and ReLU activations. The $\sigma^2$ also have 2 hidden layers with 32 nodes and ReLU activations, as well as an exponential output activation to ensure a strictly positive output value. During training, $\epsilon$-greedy exploration with $\epsilon$ starting at 0.1 and decaying linearly to 0 was performed. When generating the decision boundary plots, the deterministic policy was used without added stochasticity.

\section{Further experiments}

\subsection{MEC and Incomparability}
\label{sec:mec_experiment}

We first illustrate the problems arising when applying MEC with incomparable theories. Fig.~1(a) shows a set of preferences for the ``classic'' version of the trolley problem (Fig.~1(a)), with a single version of utilitarianism for which the choice-worthiness is the negative of the number of people harmed in the environment, and two versions of deontology: in the first, switching corresponds to a -1 choice-worthiness, while the second is ``boosted'' so that switching is given a -10 choice-worthiness.

\begin{figure}[htb]
    \centering
    \subtable[Preferences in the classic trolley problem.]{
        \begin{tabular}{c|c c}
             & Crash into 1 & Crash into X \\
            \hline
            Utilitarianism & -1 & -X \\
            Deontology & -1 & 0 \\
            Boosted Deontology & -10 & 0 \\
        \end{tabular}
        \label{tab:trolley_worthiness_mec}}
        
    \includegraphics[height=0.18in]{Figures/legends/nothing_switch.pdf}
    
    \vspace{-1.5mm}
    \subfigure[MEC (deontology)]{%
        \includegraphics[height=1.13in]{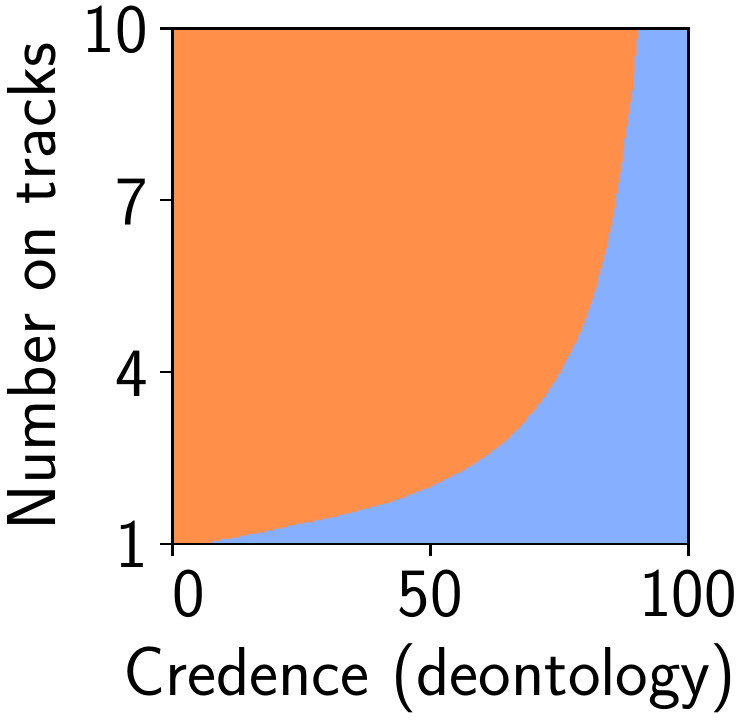}%
        \label{fig:mec_util_deontology}}%
    \subfigure[MEC (boosted deontology)]{%
        \includegraphics[height=1.13in]{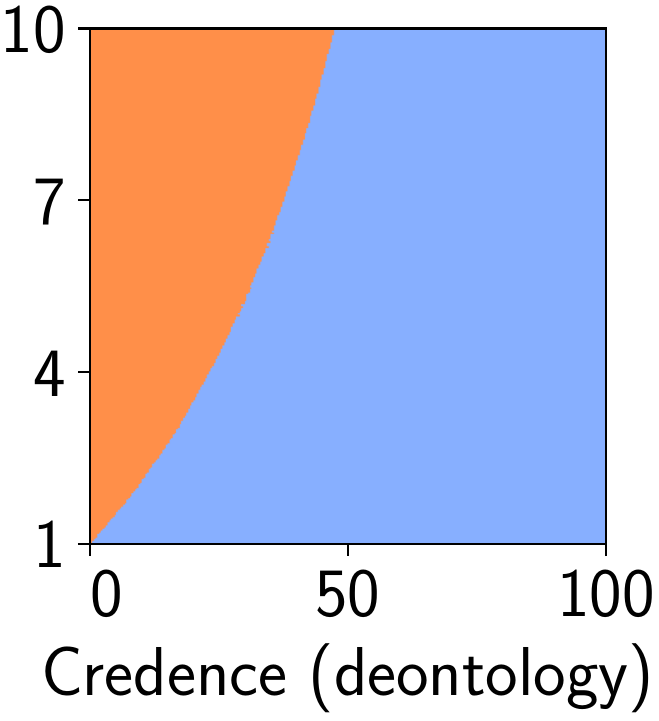}
        \label{fig:mec_util_boosted_deontology}}%
    \subfigure[Variance voting (both)]{%
        \includegraphics[height=1.13in]{Figures/new_plots/variance/classic.pdf}%
        \label{fig:variance_mec}}
    \caption{\textbf{MEC is sensitive to the particular scale of utility functions.} MEC produces inconsistent results between utilitarnism and deontology depending on whether the deontological theory is scaled by a factor of 1 (a) or 10 (b). However, because there is not truth of the matter as to how the scale of units used by the deontological theory compare to that of the utiliarian theory, rescaling should have no impact on the final results. Variance voting produces the same result no matter the rescaling (c).}
    \label{fig:mec_incomparable}
\end{figure}

These two options correspond to the same underlying preference function, and because it is unclear how to compare the units used by utilitarianism (number of people harmed in this case) and those used by deontology (in this case some measure of how much the agent caused the harms that did occur to happen), there is no fact of the matter as to which of the two representations for deontology is more correct in this context. However, as SI Fig.~\ref{fig:mec_util_deontology} and \ref{fig:mec_util_boosted_deontology} demonstrate, the choice that is made has a strong impact on the final outcome. By contrast, variance voting produces the same result no matter which scale is used for deontology (SI Fig.~\ref{fig:variance_mec}).

\subsection{Q-Learning and the Illusion of Control}
\label{sec:optimism}

It has been recognized in the multi-agent literature~\cite{russell2003q} that aggregating preferences obtained using Q-learning produces a phenomenon known as the \emph{illusion of control}: the Q-learning target $y_i = r + \gamma_i \max_{a'} Q_i(s', a')$ implicitly assumes that the action that would be taken by the policy in the next state would be whichever maximizes the reward for theory $i$. However, the preferred next state action might vary across different theories, and thus which one is ultimately taken will depend on the relative credences of the theories. 

\begin{figure}[htb]
    \centering
    \subtable[Preferences in the guard trolley problem.]{
        \begin{tabular}{c|c c c}
             & Lie to the guard & Push L & Crash into X \\
            \hline
            Util. & 0 & -1 & -X \\
            Deont. & -0.5 & -4 & 0 \\
        \end{tabular}
        \label{tab:trolley_worthiness_optimism}}

    \includegraphics[height=0.18in]{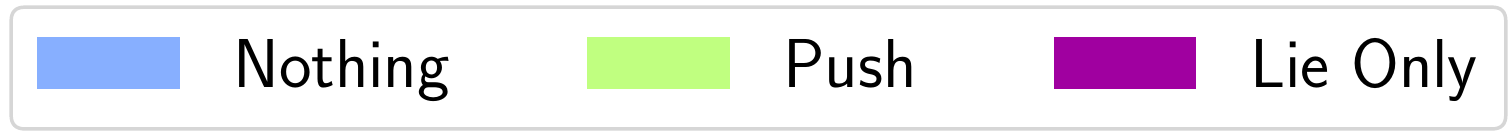}
    
    \vspace{-1.5mm}
    \subfigure[Q-learning]{%
        \includegraphics[height=1.13in]{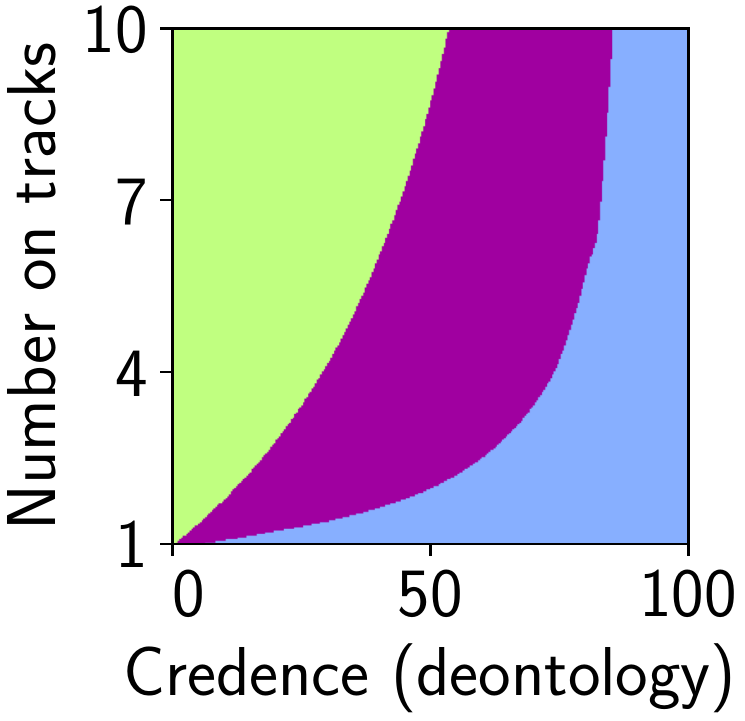}%
        \label{fig:optimism_deepq}}%
    \subfigure[Variance-SARSA]{%
        \includegraphics[height=1.13in]{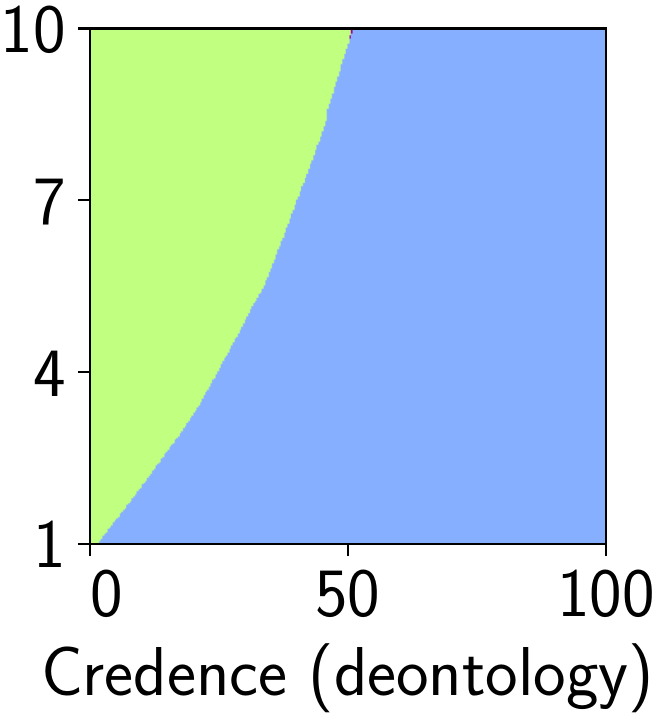}%
        \label{fig:optimism_sarsa}}%
    \subfigure[Nash voting]{%
        \includegraphics[height=1.13in]{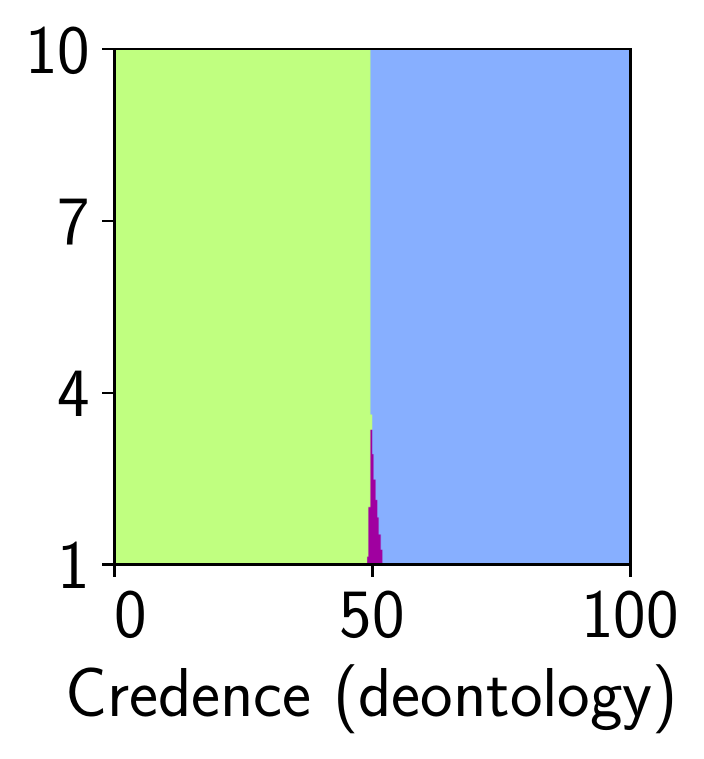}%
        \label{fig:optimism_nash}}
    \caption{\textbf{Q-learning suffers from the illusion of control.} (a) The value function is learned with Q-learning for each theory, resulting in the undesirable outcome where the ``Lie only'' option is often selected. (b) The use of Variance-SARSA results in ``Lie only'' never being chosen. (c) Nash voting is also largely immune to this outcome. ``Lie only'' is selected in rare cases in both (b) and (c), but this results from minor training instabilities rather than it being the optimal outcome for these methods (SI~\ref{app:unsable}).}
\end{figure}

This issue can be illustrated using the guard trolley problem (Fig.~1(c)): in this problem, the agent is given the option to push a large man to save the people on the tracks, but to do so it must first tell a lie to a guard protecting the large man. As can be seen in SI Tab.~\ref{tab:trolley_worthiness_optimism}, utilitarianism is indifferent to lying to the guard, while deontology views it as negative, but not nearly as bad as pushing the large man. As a result, the possibility of lying to the guard only to fail to push the large man is strictly dominated as it satisfies neither utilitarianism nor deontology, while the options of doing nothing at all or both lying the guard and pushing the large man are both possible depending on the stakes and credences involved.

As seen in SI Fig.~\ref{fig:optimism_deepq}, however, when the preferences of the different theories are trained using traditional Q-learning instead of Variance-SARSA, lying to the guard without pushing the large man is the outcome in many cases. This is because in the first step, utilitarianism's vote for lying is excessively high as the Q function for utilitarianism mistakenly believes it will be able to push the large man in the following step. Instead, unless the credence of utilitarianism is particularly high, it will get outvoted in step 2. SI Fig.~\ref{fig:optimism_sarsa} shows that when using Variance-SARSA, this illusion of control does not occur and the ``Lie Only'' option is never chosen. Like Variance-SARSA, Nash voting is immune to the illusion of control (SI Fig.~\ref{fig:optimism_nash}).

Both Variance-SARSA and Nash voting suffer from mild stability issues (SI~\ref{app:unsable}) in this particular problem, due to the fact that that the votes near the decision boundary need to be resolved with high precision to avoid the dominated ``lie only'' outcome, which is not perfectly achieved by the hyperparameters used in these experiments.

\section{Removal of irrelevant alternatives}
\label{app:doomsday_solution}

It may be objected that a trivial solution to ``doomsday'' problem introduced in Sec.~5.2 exists: it is obvious based from the preferences of the agents that ``doomsday'' is a strictly dominated option, and could therefore be taken out of the voting entirely before computing variances. However, this objection does not address the general issue, as adding a third theory with a preference for doomsday with even a small credence would force us to bring doomsday back as an option and thus the issue would come back. It may be possible to find alternate ways to avoid taking the doomsday option (or other irrelevant alternatives) into account when computing the variances (such as by only considering actions actually taken by the current policy), a possibility which we leave to future work. Another solution would be to use Nash voting, which is indeed immune to this issue, but suffers from the No Compromise and Stakes Insensitivity issues mentioned in Sections~5.1 and~5.2. A full solution to this issue would require further work, and may not be possible without producing other undesirable side-effects due to Arrow's theorem.

\section{Convergence of Variance-Sarsa and Outline of Variance-PG}

\begin{figure}[tbh]
    \centering
    
    \begin{tikzpicture}[>=latex,node distance=0.7cm and 3cm,state/.style={circle, draw=black, fill=white, thick, minimum size=5mm, on grid},terminal/.style={on grid}]
    \node[state] (A) {$s_0$};
    \node[state] (B) [above right=1.3cm and 3cm of A] {$s_1$};
    \node[state] (C) [below right=1.3cm and 3cm of A] {$s_2$};
    \node[terminal] (a) [above right=of B] {X};
    \node[terminal] (b) [below right=of B] {X};
    \node[terminal] (c) [above right=of C] {X};
    \node[terminal] (d) [below right=of C] {X};
    
    \draw[->] (A) -- node[above=1mm] {$a_0$ (0,0)$\:\:$} ++ (B);
    \draw[->] (A) -- node[midway,below=1mm] {$a_1$ (0,0)$\:\:$} ++ (C);
    \draw[->] (B) -- node[above=1mm] {$a_0$ (0,100)} ++ (a);
    \draw[->] (B) -- node[below=2mm] {$a_1$ (-4,80)} ++ (b);
    \draw[->] (C) -- node[above=1mm] {$a_0$ (100,0)} ++ (c);
    \draw[->] (C) -- node[below=2mm] {$a_1$ (80,-4)} ++ (d);
    \end{tikzpicture}%
    
    \caption{\textbf{An MDP in which Variance-Sarsa cannot converge.} The X's correspond to terminal states. Above each arrow is the action ($a_0$ or $a_1$) followed by the choice-worthiness of taking the given action in that state according to each theory (e.g. the choice worthiness of taking $a_0$ in $s_1$ is 0 according to Theory 1 and 100 according to Theory 2).}
    \label{fig:variance_sarsa_conv}
\end{figure}
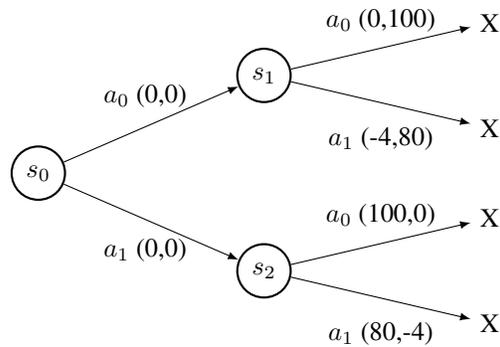

Although Variance-Sarsa converges in all the examples presented in this work, through exploiting the determinisic nature of Variance-Sarsa's policy it is possible to construct pathological examples in which convergence is impossible.

Fig.~\ref{fig:variance_sarsa_conv} shows such an example. We will suppose that Theory 1 and Theory 2 both have a credence of 0.5. Since $a_0$ is dominant in both $s_1$ and $s_2$, if we assume no discounting we have $Q_1(s_0, a_0) = 0$, $Q_1(s_0, a_1) = 100$, $Q_2(s_0, a_0) = 100$ and $Q_2(s_0, a_1) = 0$, while the Q-values at $s_1$ and $s_2$ are simply the direct choice-worthiness values at those states. 

Now suppose that at a particular time, the deterministic policy found by Variance-Sarsa is to choose $a_0$ at $s_0$. Then the $\sigma_i^2$ will be the average of the variances of the Q-values at $s_0$ and $s_1$ since those two states are visited equally often, and $s_2$ is never visited. We can thus calculate that $\sigma_1^2 = 1252$ and $\sigma_2^2 = 1300$.

Assuming $\varepsilon = 0$ for simplicity, calculating the votes for the action to take at $s_0$ gives a total vote of approximately -0.01317 in favor of $a_0$ and 0.01317 in favor of $a_1$ (intuitively, the magnitudes of the Q-values are equal at $s_0$ but Theory 1 has lower variance, so its preferred action prevails). Thus, the variances produced when a deterministic policy chooses $a_0$ at $s_0$ require a change of policy to choose $a_1$ instead at $s_0$. The symmetric nature of the MDP, however, implies that choosing $a_1$ in $s_0$ will produce variances that will favor switching to choosing $a_0$ instead. Hence, an algorithm producing deterministic policies such as Variance-Sarsa will cycle between choosing $a_0$ and $a_1$ at $s_0$, and thus never converge in this example MDP. Note that such \emph{cycling} is a common pathology in multi-agent RL (MARL), and that Variance-Sarsa is in effect a MARL algorithm (e.g.\ the dynamics of $\sigma^2_i$ can be seen to arise from interactions between the policies of each theory; see also Sec.~4.3 where we show that variance voting arises from Nash voting under some constraints).

A solution to this particular example would be an algorithm capable of reaching a stochastic policy. In particular, a policy which chooses action $a_0$ and $a_1$ $50\%$ of the time each would be at equilibrium in the example above. We hypothesize that such a policy could be trained using an actor-critic policy gradient algorithm~\cite{sutton1998reinforcement} in which $Q_i$ and $\sigma_i^2$ are learned in the same way as Variance-Sarsa but where the Variance-Sarsa vote replaces the action value in the policy gradient update, i.e. the policy $\pi_{\theta}(a|s)$ would be updated in the direction $\left(\sum_i C_i \frac{Q_i(s, a) - \mu_i(s)}{\sqrt{\sigma_i^2} + \varepsilon}\right)\nabla_\theta \log \pi_\theta(a|s)$. We call this possible algorithm Variance-PG and hypothesize that it would always converge to a stable equilibrium under the assumption of perfect training.

\section{Quadratic Cost in Nash Voting}
\label{app:quadratic_nash}

\begin{figure}[tbh]
    \centering

    \subfigure[Double trolley problem (Fig.~3(b)).]{%
        \includegraphics[height=1.13in]{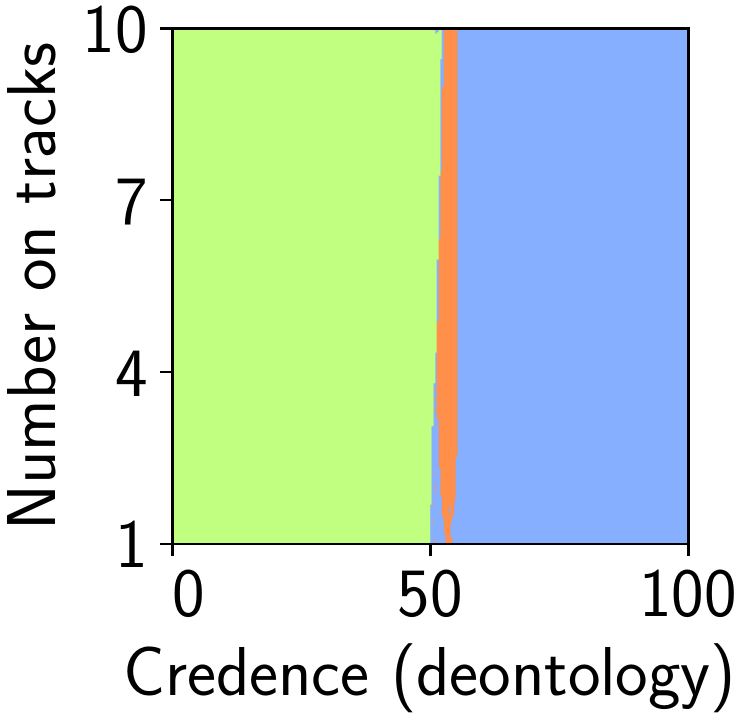}%
        \label{fig:compromise_nash2}}%
    \subfigure[Double trolley problem with unknown adversary (Fig.~3(c)).]{%
        \includegraphics[height=1.13in]{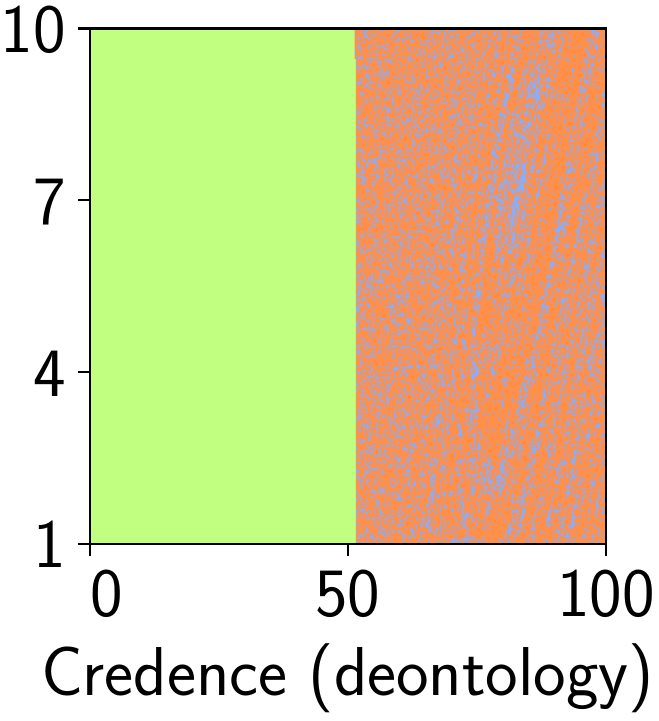}%
        \label{fig:compromise_nash2_unknown}}%
    \subfigure[Guard trolley problem (SI Fig.~\ref{fig:optimism_nash}).]{%
        \includegraphics[height=1.13in]{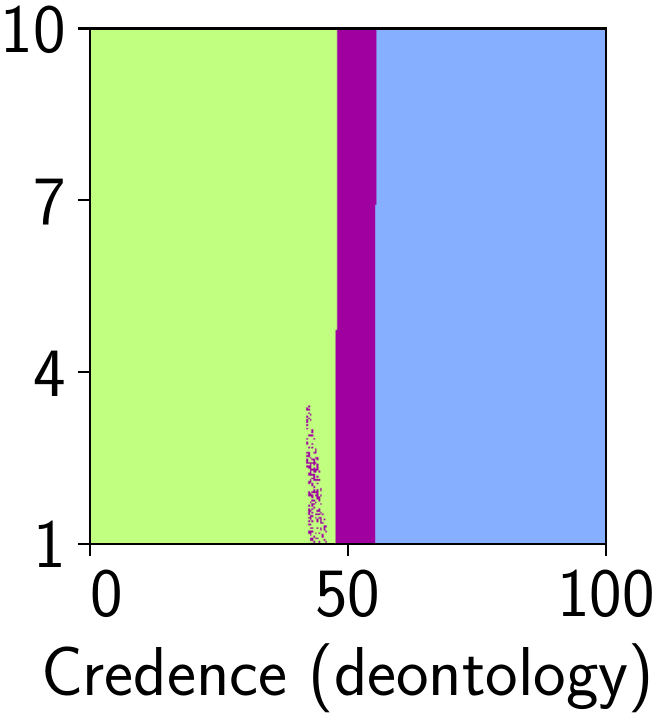}%
        \label{fig:optimism_nash2}}%
    \caption{\textbf{Experiments with significant differences between Nash voting with quadratic cost and Nash voting with absolute value cost.} Each subfigure corresponds to a particular Nash voting experiment from the main paper, but uses a quadratic cost instead of an absolute value cost.}
    \label{fig:nash2_differences}
\end{figure}

The results in Fig.~2(b), 2(c) and 4(d) are identical whether an absolute value cost or a quadratic cost is used. SI Fig.~\ref{fig:nash2_differences} shows the three cases in which different results are obtained with a quadratic cost function. In SI Fig.~\ref{fig:compromise_nash2}, the compromise solution is produced near the decision  boundary, which might indicate that Nash voting with a quadratic cost does not suffer from No Compromise. However, the high instability of Nash voting in this particular problem (SI Fig.~\ref{fig:compromise_nash2_unstable}) as well as the fact that Nash voting with quadratic cost produces the strictly dominated ``Lie Only'' solution in the guard trolley problem (SI Fig.~\ref{fig:optimism_nash2}) suggest that this apparently positive outcome is likely due to training instabilities rather than genuine robustness to No Compromise. In SI Fig.~\ref{fig:compromise_nash2_unknown}, the solution to the double trolley problem with unknown adversary has an extremely poorly defined decision boundary. All examples are highly unstable, as shown in SI Fig.~\ref{fig:compromise_nash2_unstable}, \ref{fig:compromise_nash2_unknown_unstable} and \ref{fig:optimism_nash2_unstable}. 

\section{Related work in philosophy and machine ethics}
\label{app:related_phil}

The question of moral uncertainty has only recently been of focused interest within moral philosophy~\cite{macaskill2014normative,lockhart2000moral,sepielli2013moral}, and this work is among the first to connect it to machine learning~\cite{bogosian2017implementation}, and the first to offer a concrete implementation. 

One hope is that working towards concrete implementations of moral uncertainty may also offer philosophical insights, as computational implementation requires notions to become workably precise -- in the words of Daniel Dennett, AI ``makes philosophy honest~\cite{dennett2006computers}.'' 
In this way, this paper relates to computational philosophy~\cite{sep-computational-philosophy,thagard1993computational}, wherein computational techniques are used to advance philosophy. For example, agent-based works that study the evolution of cooperation~\cite{nitschke2005emergence} or that model artificial morality~\cite{danielson2002artificial}, have connections to the study of ethics. Differing from those works, our approach focuses on integrating multiple human ethical views into a single agent.

This paper can be seen as fitting into the field of machine ethics~\cite{allen2006machine,wallach2008moral}, which seeks to give moral capabilities to computational agents. Many approaches in machine ethics seek to create an agent that embodies a \emph{particular} moral theory -- e.g.\ agents that implement versions of utilitarianism~\cite{anderson2005towards}, deontology~\cite{hooker2018toward}, and prima facie duties. Our work complements such approaches by seeking to combine multiple ethical theories within an agent. Additionally, like \citet{abel2016reinforcement}, we attempt to highlight practical bridges between machine ethics and RL.

\section{Instability}
\label{app:unsable}

We observe empirically that the decision boundary does not always reach a stable equilibrium during training. Thus, in the case of unstable experiments, the decision boundary plot that (according to a subjective assessment) best represents the equilibrium point was used in the main text instead of the decision plot produced at the end of training. This SI section provides the full sets of 20 decision plots for all unstable experiments.

The instability phenomenon is most common in Nash voting (SI Fig.~\ref{fig:optimism_nash_unstable}, \ref{fig:stakes_nash_sequential_unstable} \ref{fig:compromise_nash_unknown_unstable} and \ref{fig:compromise_nash_sequential_unstable}), though it occasionally occurs in variance voting (SI Fig.~\ref{fig:optimism_sarsa_unstable}). Further, SI Fig.~\ref{fig:compromise_nash_sequential_unstable} and \ref{fig:compromise_nash2_sequential_unstable} show the outcome of an experiment not included in the main text in which iterated Nash voting (presented in Sec.~5.1) is used on the double trolley problem (Sec.~5.2). In this experiment, Nash voting is completely unstable and it is unclear whether a stable equilibrium exists at all. By contrast, the only unstable case in variance voting (SI Fig.~\ref{fig:optimism_sarsa_unstable}) may be alleviated by tweaking the hyperparameters (for example by annealing the learning rate to ensure convergence), since it merely oscillates around an equilibrium instead of converging to it. 

\newcommand{\unstableTable}[1]{
\subfigure[0.5M iterations]{
\includegraphics[height=0.17\linewidth]{Figures/grouped/#1/0000500000.pdf}}
\subfigure[1.0M iterations]{
\includegraphics[height=0.17\linewidth]{Figures/grouped/#1/0001000000.pdf}}
\subfigure[1.5M iterations]{
\includegraphics[height=0.17\linewidth]{Figures/grouped/#1/0001500000.pdf}}
\subfigure[2.0M iterations]{
\includegraphics[height=0.17\linewidth]{Figures/grouped/#1/0002000000.pdf}}
\subfigure[2.5M iterations]{
\includegraphics[height=0.17\linewidth]{Figures/grouped/#1/0002500000.pdf}}
\subfigure[3.0M iterations]{
\includegraphics[height=0.17\linewidth]{Figures/grouped/#1/0003000000.pdf}}
\subfigure[3.5M iterations]{
\includegraphics[height=0.17\linewidth]{Figures/grouped/#1/0003500000.pdf}}
\subfigure[4.0M iterations]{
\includegraphics[height=0.17\linewidth]{Figures/grouped/#1/0004000000.pdf}}
\subfigure[4.5M iterations]{
\includegraphics[height=0.17\linewidth]{Figures/grouped/#1/0004500000.pdf}}
\subfigure[5.0M iterations]{
\includegraphics[height=0.17\linewidth]{Figures/grouped/#1/0005000000.pdf}}
\subfigure[5.5M iterations]{
\includegraphics[height=0.17\linewidth]{Figures/grouped/#1/0005500000.pdf}}
\subfigure[6.0M iterations]{
\includegraphics[height=0.17\linewidth]{Figures/grouped/#1/0006000000.pdf}}
\subfigure[6.5M iterations]{
\includegraphics[height=0.17\linewidth]{Figures/grouped/#1/0006500000.pdf}}
\subfigure[7.0M iterations]{
\includegraphics[height=0.17\linewidth]{Figures/grouped/#1/0007000000.pdf}}
\subfigure[7.5M iterations]{
\includegraphics[height=0.17\linewidth]{Figures/grouped/#1/0007500000.pdf}}
\subfigure[8.0M iterations]{
\includegraphics[height=0.17\linewidth]{Figures/grouped/#1/0008000000.pdf}}
\subfigure[8.5M iterations]{
\includegraphics[height=0.17\linewidth]{Figures/grouped/#1/0008500000.pdf}}
\subfigure[9.0M iterations]{
\includegraphics[height=0.17\linewidth]{Figures/grouped/#1/0009000000.pdf}}
\subfigure[9.5M iterations]{
\includegraphics[height=0.17\linewidth]{Figures/grouped/#1/0009500000.pdf}}
\subfigure[10.0M iterations]{
\includegraphics[height=0.17\linewidth]{Figures/grouped/#1/0010000000.pdf}}
}

\clearpage
\begin{figure*}[htbp!]
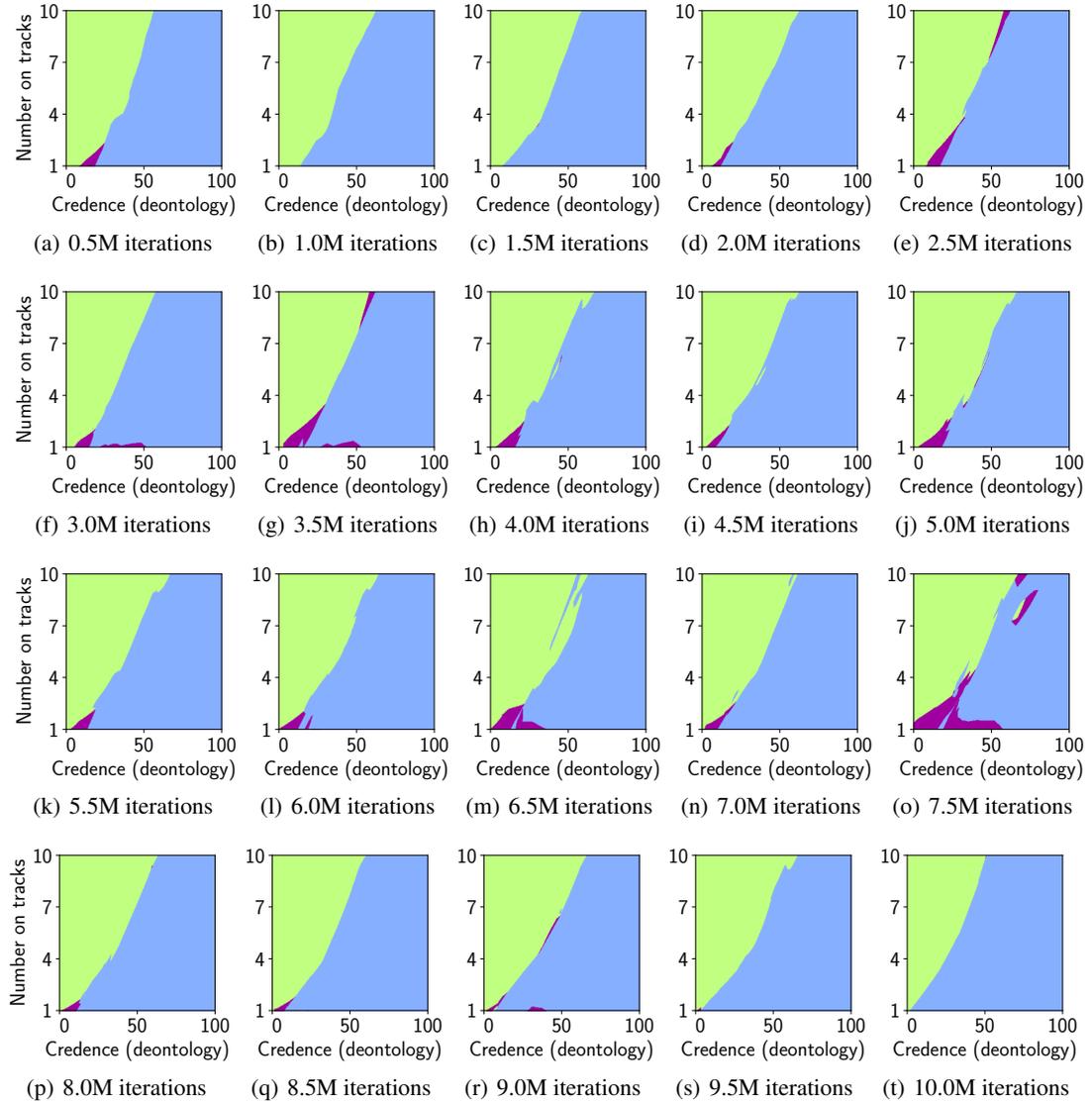

    \centering
\unstableTable{optimism/sarsa}
\caption{\textbf{Instability of variance voting in the guard trolley problem} (SI Fig.~\ref{fig:optimism_sarsa}).}
\label{fig:optimism_sarsa_unstable}
\end{figure*}

\clearpage

\begin{figure*}[!hbtp]
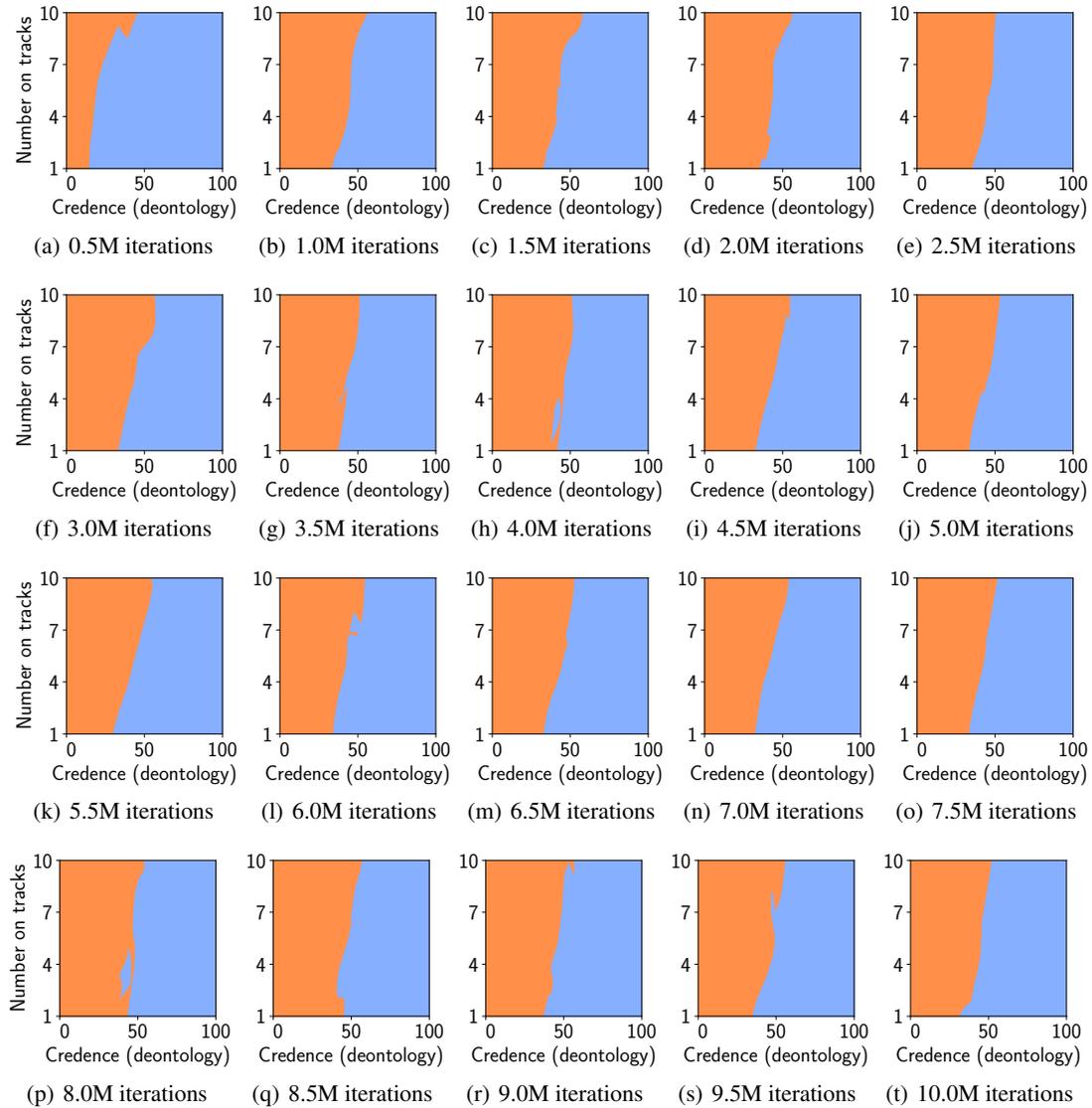

\centering
\unstableTable{stakes/nash_sequential}
\caption{\textbf{Instability of iterated Nash voting} (Fig.~2(c)).}
\label{fig:stakes_nash_sequential_unstable}
\end{figure*}

\begin{figure*}[hbtp]
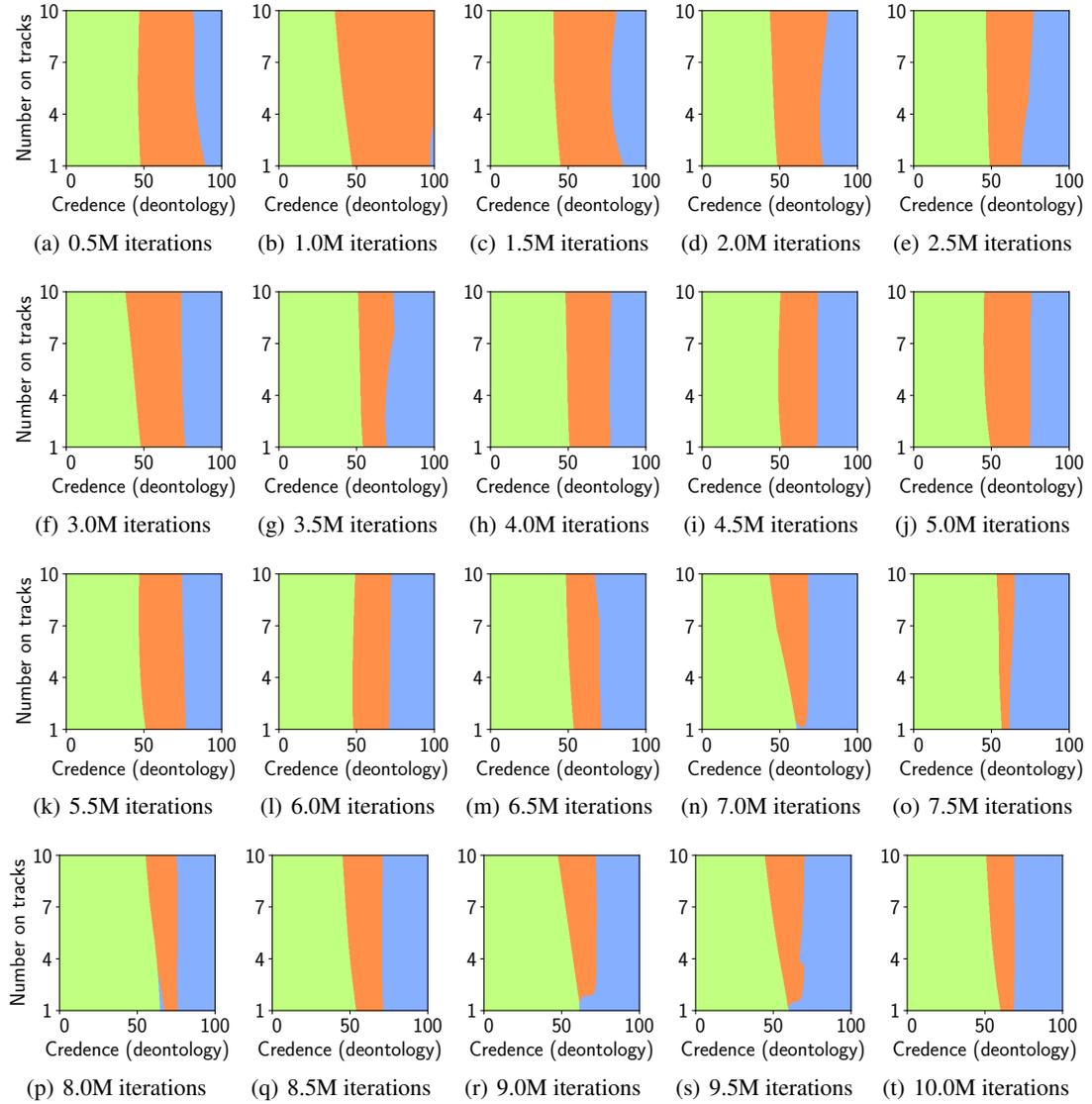

\centering
\unstableTable{compromise/nash_unknown}
\caption{\textbf{Instability of Nash voting with unknown adversary} (Fig.~3(c)).}
\label{fig:compromise_nash_unknown_unstable}
\end{figure*}

\begin{figure*}[hbtp]
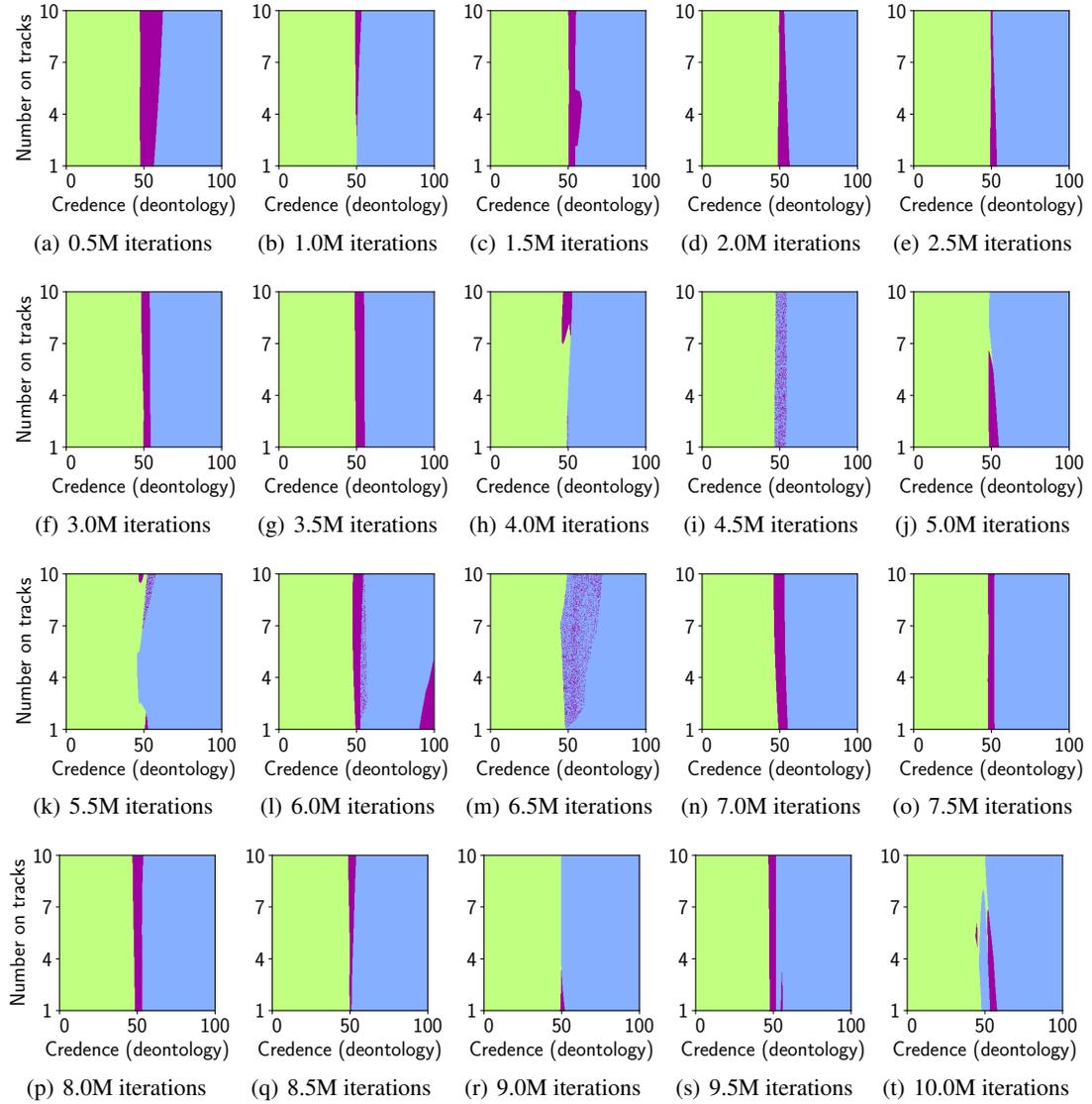

    \centering
\unstableTable{optimism/nash}
\caption{\textbf{Instability of Nash voting in the guard trolley problem} (SI Fig.~\ref{fig:optimism_nash}).}
\label{fig:optimism_nash_unstable}
\end{figure*}

\begin{figure*}[hbtp]
    \centering
\unstableTable{compromise/nash_sequential}
\caption{\textbf{Instability of iterated Nash voting in the double trolley problem}.}
\label{fig:compromise_nash_sequential_unstable}
\end{figure*}

\clearpage

\begin{figure*}[!hbtp]
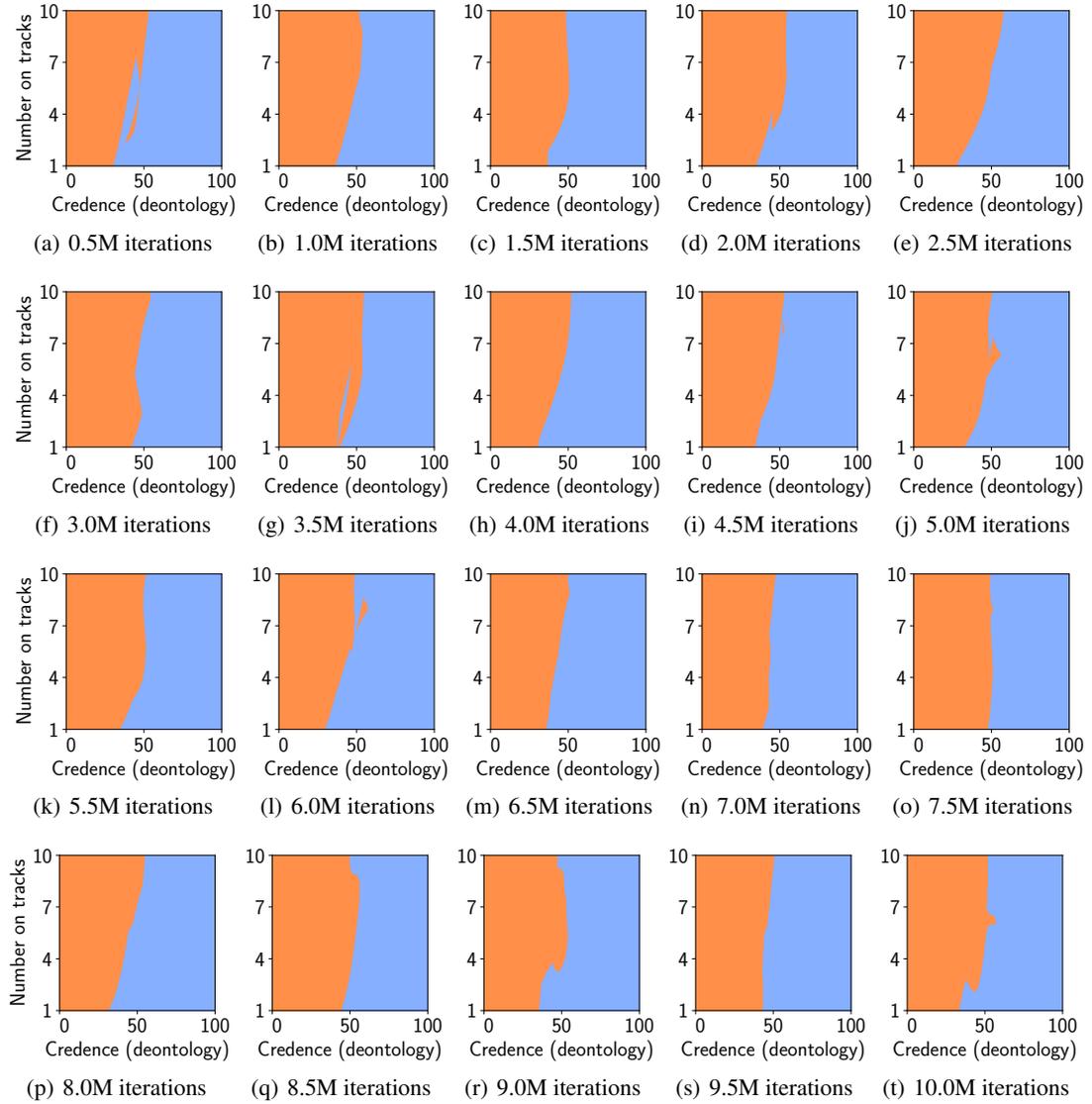

\centering
\unstableTable{stakes/nash2_sequential}
\caption{\textbf{Instability of iterated Nash voting with quadratic cost}.}
\label{fig:stakes_nash2_sequential_unstable}
\end{figure*}

\begin{figure*}[hbtp]
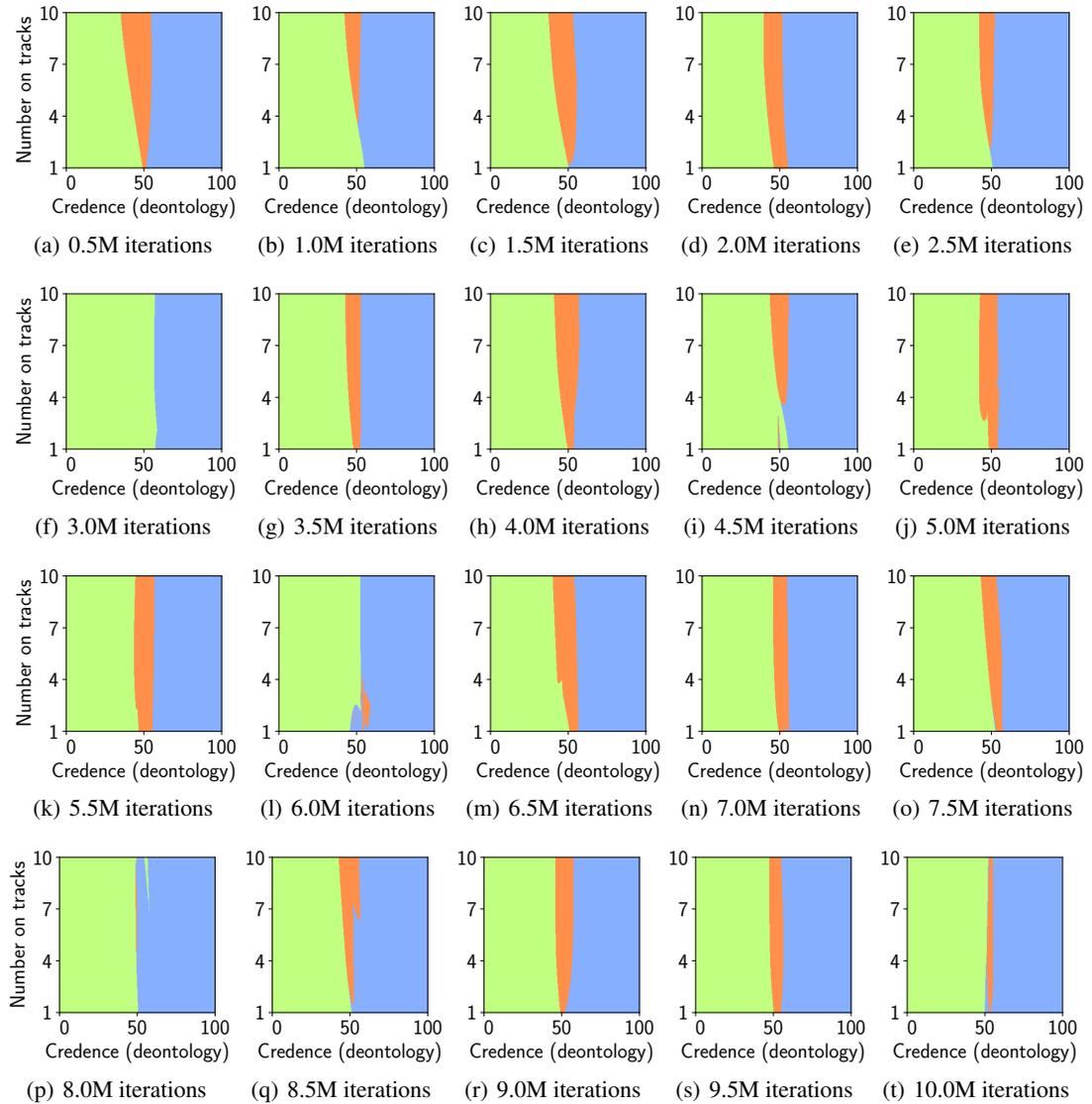

\centering
\unstableTable{compromise/nash2}
\caption{\textbf{Instability of Nash voting with quadratic cost} (SI Fig.~\ref{fig:compromise_nash2}).}
\label{fig:compromise_nash2_unstable}
\end{figure*}

\begin{figure*}[hbtp]
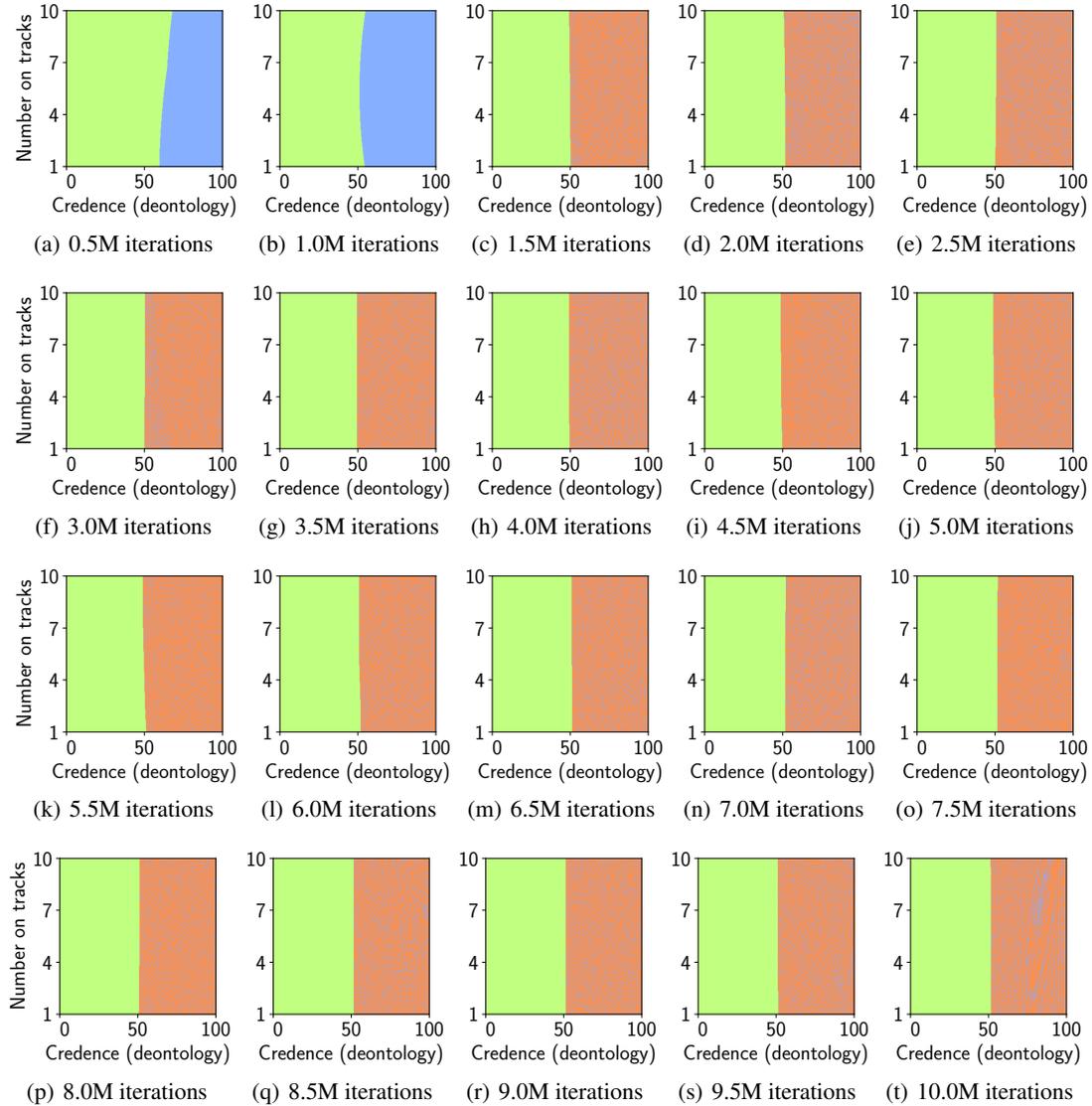

\centering
\unstableTable{compromise/nash2_unknown}
\caption{\textbf{Instability of Nash voting with unknown adversary with quadratic cost} (SI Fig.~\ref{fig:compromise_nash2_unknown}).}
\label{fig:compromise_nash2_unknown_unstable}
\end{figure*}

\begin{figure*}[hbtp]
    \centering
\unstableTable{compromise/nash2_sequential}
\caption{\textbf{Instability of iterated Nash voting in the double trolley problem with quadratic cost}.}
\label{fig:compromise_nash2_sequential_unstable}
\end{figure*}

\begin{figure*}[hbtp]
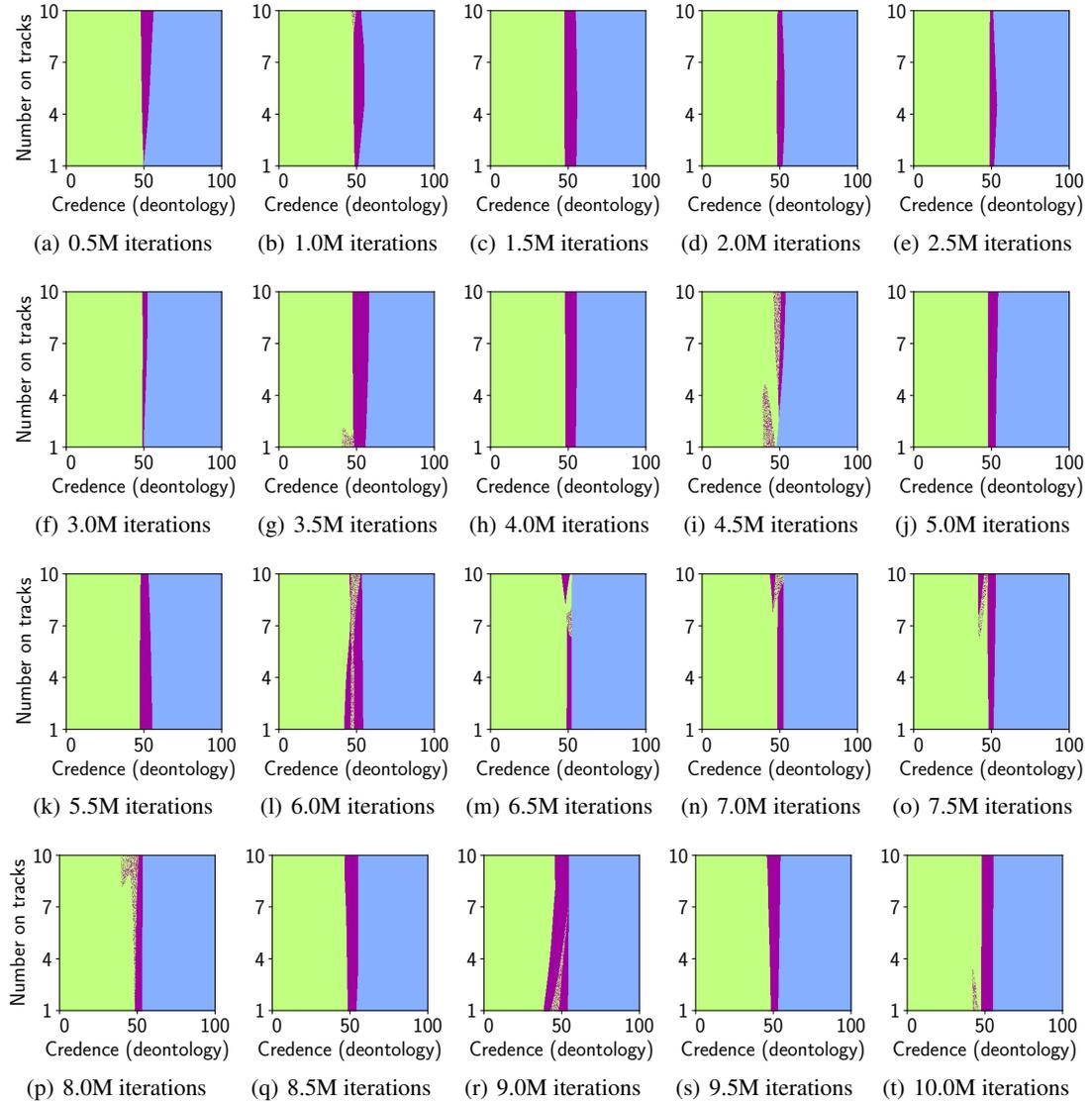

    \centering
\unstableTable{optimism/nash2}
\caption{\textbf{Instability of Nash voting with quadratic cost in the guard trolley problem} (SI Fig.~\ref{fig:optimism_nash2}).}
\label{fig:optimism_nash2_unstable}
\end{figure*}

\end{document}